%% file: main.tex
\documentclass{article}

\usepackage{microtype}
\usepackage{graphicx}
\usepackage{subcaption}
\usepackage{booktabs} 

\usepackage{hyperref}

\usepackage[accepted]{icml2026}

\usepackage{amsmath}
\usepackage{amssymb}
\usepackage{mathtools}
\usepackage{amsthm}

\usepackage{url}
\usepackage{xspace}
\usepackage{graphicx}
\usepackage{threeparttable}
\usepackage{booktabs}
\usepackage{pifont}
\usepackage{float}
\usepackage{booktabs}
\usepackage{footmisc}
\usepackage{wrapfig}
\usepackage[normalem]{ulem}
\usepackage{enumitem}

\usepackage[capitalize,noabbrev]{cleveref}

\theoremstyle{plain}

\theoremstyle{definition}

\theoremstyle{remark}

\usepackage[textsize=tiny]{todonotes}

\input{math_commands.tex}

\usepackage{ml-defs}

\usepackage[utf8]{inputenc} 
\usepackage{newunicodechar}
\newunicodechar{ }{\,}
\DeclareUnicodeCharacter{0308}{\"}
\usepackage[T1]{fontenc}    
\usepackage{url}            
\usepackage{booktabs}       
\usepackage{amsfonts}       
\usepackage{nicefrac}       
\usepackage{microtype}      
\usepackage{xcolor}         
\usepackage{acronym}  
\usepackage{bm}
\usepackage{threeparttable}
\usepackage{graphicx}
\usepackage[normalem]{ulem}

\usepackage{siunitx}
\usepackage{stfloats}   

\definecolor{linkcol}{HTML}{3073AD}  
\definecolor{citecol}{HTML}{3073AD} 
\definecolor{urlcol}{HTML}{3073AD} 
\hypersetup{
   colorlinks=true,
   linkcolor=linkcol,
   citecolor=citecol,
   urlcolor=urlcol,
}

\newcommand{\stoptocwriting}{%
   \addtocontents{toc}{\protect\setcounter{tocdepth}{-5}}}

\newcommand{\deleted}[1]{{}}

\acrodef{ai}[AI]{artificial intelligence}
\acrodef{ml}[ML]{machine learning}
\acrodef{cadd}[CADD]{computer-aided drug design}
\acrodef{lbdd}[LBDD]{ligand-based drug design}
\acrodef{sbdd}[SBDD]{structure-based drug design}
\acrodef{smiles}[SMILES]{Simplified Molecular Input Line Entry System}
\acrodef{pdb}[PDB]{Protein Data Bank}
\acrodef{qsar}[QSAR]{quantitative structure-activity relationship}
\acrodef{pli}[PLI]{protein-ligand interaction}
\acrodef{vs}[VS]{virtual screening}
\acrodef{vnegnn}[VN-EGNN]{equivariant graph neural network with virtual nodes}
\acrodef{egnn}[EGNN]{equivariant graph neural network}
\acrodef{mlp}[MLP]{multi-layer-perceptron}
\acrodef{gnn}[GNN]{graph neural network}

\icmltitlerunning{Contrastive Geometric Learning Unlocks Unified Structure- and Ligand-Based Drug Design}

\begin{document}

\twocolumn[
  \icmltitle{Contrastive Geometric Learning Unlocks Unified \\Structure- and Ligand-Based Drug Design}



  \icmlsetsymbol{equal}{*}

  \begin{icmlauthorlist}
    \icmlauthor{Lisa Schneckenreiter}{jku}
    \icmlauthor{Sohvi Luukkonen}{jku}
    \icmlauthor{Lukas Friedrich}{merck}
    \icmlauthor{Daniel Kuhn}{merck}
    \icmlauthor{Günter Klambauer}{jku,med}
  \end{icmlauthorlist}

  \icmlaffiliation{jku}{ELLIS Unit Linz, LIT AI Lab, Institute for Machine Learning, Johannes Kepler University Linz, Austria}
  \icmlaffiliation{merck}{Medicinal Chemistry \& Drug Design, Merck Healthcare KGaA, Darmstadt, Germany}
  \icmlaffiliation{med}{Clinical Research Institute for Medical Artificial Intelligence, Johannes Kepler University, Linz, Austria}

  \icmlcorrespondingauthor{Lisa Schneckenreiter}{schneckenreiter@ml.jku.at}

  \icmlkeywords{Machine Learning, ICML}

  \vskip 0.3in
]



\printAffiliationsAndNotice{}  

\begin{abstract}
    Structure-based and ligand-based computational drug design have traditionally relied on disjoint data sources and modeling assumptions, limiting their joint use at scale.
    In this work, we introduce \textbf{Con}trastive \textbf{G}eometric \textbf{L}earning for \textbf{U}nified Computational \textbf{D}rug D\textbf{e}sign (ConGLUDe), a single contrastive geometric model that unifies structure- and ligand-based training. 
    ConGLUDe couples a geometric protein encoder that produces whole-protein representations and implicit embeddings of predicted binding sites with a fast ligand encoder, removing the need for predefined pockets. 
    By aligning ligands with both global protein representations and multiple candidate binding sites through contrastive learning, ConGLUDe supports ligand-conditioned pocket prediction in addition to virtual screening and target fishing, while being trained jointly on protein–ligand complexes and large-scale bioactivity data.
    Across diverse benchmarks, ConGLUDe achieves competitive zero-shot virtual screening performance, substantially outperforms existing methods on a challenging target fishing task, and demonstrates state-of-the-art ligand-conditioned pocket selection. 
    These results highlight the advantages of unified structure–ligand training and position ConGLUDe as a step toward general-purpose foundation models for drug discovery.
    
\end{abstract}

\stoptocwriting

\section{Introduction}
\label{introduction}

\textbf{Protein–ligand interactions play a crucial role in drug discovery.}  
Most therapeutics are small molecules, referred to as ligands, that bind to disease-associated protein targets to modulate their function \citep{kinch20242023}.
Understanding \acp{pli} through modeling atomic interactions at binding sites or more general ligand bioactivity measures enables rational drug design \citep{gohlke2000knowledge,du2016insights}. 
Computational methods for predicting and analyzing these interactions have been used for decades, and recent advances in \ac{ai} and \ac{ml} have greatly expanded their accuracy and applicability. 
These approaches are commonly grouped into two paradigms: \ac{sbdd} and \ac{lbdd} \citep{macalino2015role,vemula2023cadd}, with disjoint data sources and modalities.
\begin{figure}[t]
  \centering
  \includegraphics[width=0.49\textwidth]{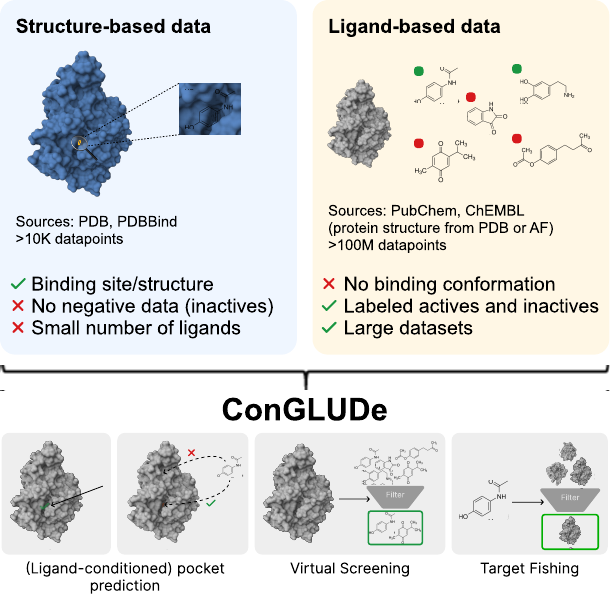}
  \caption{Joint training of ConGLUDe on structure- and ligand-based data enables (ligand-conditioned) pocket prediction, virtual screening, and target fishing.}
  \label{fig:overview}
\end{figure}

\textbf{\Ac{sbdd} relies on structural data describing the 3D conformations of ligands bound to protein binding sites.}
Generally, this information is derived from experimentally resolved protein-ligand complexes, obtained using, e.g., X-ray crystallography or NMR spectroscopy \citep{mutharasappan2020experimental}, a technically demanding process that has historically limited the application of SBDD to only a fraction of known proteins. Experimental structures are systematically archived in the \ac{pdb} \citep{berman2000protein}, which contains $\sim$245k entries\footnote{From https://www.rcsb.org/stats/growth/growth-released-structures. Accessed on 15/12/2025.}. This provides a valuable resource for \ac{sbdd}, although only a fraction of entries contain biologically relevant ligands, and the total number of available ligands is limited.

In computational SBDD, the binding potential of candidate molecules to a specific pocket is typically assessed using methods such as docking \citep{kuntz1982geometric, fan2019progress}, molecular dynamics \citep{devivo2016role}, or free energy perturbation \citep{beveridge1989free,cournia2021free}. More recently, AI innovations have enabled the holistic prediction of protein-ligand complexes -- without explicit binding pocket information as input -- either with AI-based blind docking methods \citep{stark2022equibind,corso2023diffdock,pei2023fabind}, or foundation models for molecular structure prediction of biological complexes \citep{abramson2024accurate,wohlwend2024boltz,boitreaud2024chai}. However, by modeling exact binding positions, these methods are inherently computationally demanding, making them unsuitable for large-scale virtual screening.

\textbf{\Ac{lbdd} uses bioactivity data from large-scale biochemical assays.} 
Rather than modeling explicit binding geometries, \ac{lbdd} learns ligand activity directly from experimental assay outcomes and therefore does not depend on knowledge of the ligand’s binding conformation within the protein target \citep{merz2010drug}. This paradigm is supported by large public resources such as PubChem, which contains approximately 300M bioactivity measurements \citep{kim2025pubchem}, and ChEMBL, which provides curated bioactivity data across thousands of targets \citep{gaulton2011chembl}.

As a result, \ac{ml} has been central to \ac{lbdd} since the early 1990s \citep{hansch1962correlation,muratov2020qsar}, progressing from classical \ac{qsar} models such as support vector machines \citep{burbidge2001drug} and random forests \citep{svetnik2003random} to gradient boosting \citep{babajide2016bioactive,sheridan2016extreme} and deep learning approaches \citep{lenselink2017beyond, mayr2018large, yang2019analyzing}. While early methods required target-specific data, recent few-shot and zero-shot learning techniques extend activity prediction to data-scarce targets \citep{vella2022few,schimunek2023mhnfs,seidl2023clamp}. Complementarily, proteochemometrics enables generalization to unseen targets by incorporating protein representations derived from sequence or structure \citep{lapinsh2001development,ozturk2018deepdta,bongers_proteochemometrics_2019,svensson2024hyperpcm}. This has been supported by advances such as AlphaFold \citep{jumper2021alphafold}, which enable the extension of structure-based protein embeddings to targets without experimental PDB entries.

\textbf{Recently, contrastive learning has emerged as a powerful paradigm for modeling \acp{pli}.}
These models embed protein and ligand representations in a shared latent space, where interactions are inferred via representational similarity, yielding substantial computational efficiency that enables genome-wide screens of large-scale compound libraries \citep{jia2026drugclip}.
When trained on structure-based data, they typically rely on explicitly encoding predefined binding pockets, thereby restricting their applicability to targets with known binding sites \citep{gao2023drugclip, han2024hashing,feng2025hierarchical,wang2025hypseek,he2025s2drug}. Conversely, contrastive models trained on ligand-based data use sequence- or structure-level encodings of entire proteins, allowing broader applicability but without enabling pocket-specific predictions \citep{singh2023contrastive,mcnutt2024sprint}.

\textbf{Combining both data types could yield more informative protein and ligand representations.} 
Previous attempts at unifying structure- and ligand-based data augment PDB complexes with ligand-based data either by assigning active ligands to known binding pockets -- implicitly assuming the correct binding site \citep{gao2023drugclip, feng2025hierarchical, wang2025hypseek} --, or by (blind) docking, which is computationally expensive and introduces pose uncertainty \citep{francoeur2020crossdocked, brocidiacono2024bigbind, liu2024helixdock}. Other approaches use the two data types to supervise different components of the model, such as a co-folding-based structure prediction module with a separate affinity-prediction head \citep{passaro2025boltz2}, or separate encoders for protein sequences and binding pockets, which requires both input types at inference \citep{he2025s2drug}. Yet, no prior method fully integrates structure- and ligand-based data in an efficient way that enables pocket-level predictions without relying on predefined pockets.

\textbf{Contrastive geometric learning unlocks seamless integration of structure- and ligand-based data.}
We introduce \emph{Contrastive Geometric Learning for Unified Computational Drug Design} (ConGLUDe), a method that combines a geometric protein encoder -- providing whole-protein representations and implicit embeddings of predicted binding sites, thus eliminating the need for predefined pockets -- with a fast, efficient ligand encoder via contrastive learning. 
Integrating pocket prediction capabilities into the protein encoder enables learning from both structure- and ligand-based data without losing pocket-specific information. 
The extension of the well-established CLIP-style contrastive loss to a third axis -- along different predicted binding sites on a protein -- allows for the introduction of the novel ligand-conditioned pocket prediction task, which focuses on the identification of a specific ligand’s binding site without resorting to inefficient blind docking or co-folding.
In addition, ConGLUDe can be directly applied to a wide range of other drug discovery tasks, including virtual screening, target fishing, and binding site prediction (Figure~\ref{fig:overview}).

\textbf{Key contributions.}
\begin{itemize}[nosep]

    \item ConGLUDe is the first \textit{end-to-end} architecture to integrate classical structure-based applications, such as binding site prediction, with large-scale protein–ligand interaction modeling, performing \textit{orders of magnitude faster} than docking or co-folding.
    \item ConGLUDe enables learning from both \textit{structure-based} protein-ligand binding conformations and millions of \textit{ligand-based} bioactivity measurements.
    \item ConGLUDe excels across \textit{diverse drug discovery tasks}, achieving competitive virtual screening, substantially outperforming all baselines in target fishing, and demonstrating state-of-the-art (ligand-conditioned) pocket prediction.

\end{itemize}

\section{Background and Preliminaries}

\subsection{Notation and Definitions}

\textbf{Protein–ligand interaction data points.}
A \ac{pli} data point is defined as a triplet $(\mathcal{G}, \mathcal{M}, y)$, where $\mathcal{G}$ denotes a protein, $\mathcal{M}$ a ligand (typically a small molecule), and $y$ a binary or real-valued label.
In structure-based datasets, \acp{pli} are derived from experimentally resolved 3D structures of protein-ligand complexes. Protein–ligand pairs with observed co-crystal structures are labeled as positives ($y=1$), while all other combinations are treated as negatives ($y=0$).
In contrast, ligand-based datasets provide activity measurements for a large set of small molecules tested against a given target protein, typically obtained through biological assays. Labels may be binary (active: $y=1$, inactive: $y=0$) or continuous affinity values ($y \in \mathbb{R}$), such as $\text{IC}_{50}$ or $\text{K}_{\text{d}}$.

\textbf{Protein and ligand representations.}
We represent proteins as geometric graphs $\mathcal{G}$, where each node corresponds to an amino acid residue. Each node is assigned a 3D coordinate (specifically, the position of the C$_\alpha$ atom) and a feature vector encoding residue-specific properties, extracted using ESM-2~\citep{lin2023evolutionary}. Edges connect each node to a maximum of 10 nearest neighbors within a 10~\r{A} radius.
Ligands are represented as fixed-length vectors constructed by concatenating Morgan fingerprints~\citep{morgan1965generation} with RDKit chemical descriptors~\citep{landrum2006rdkit}.

\textbf{Definition of binding sites.}  
Structure-based datasets enable direct annotation of protein binding sites -- the regions where ligands interact with the protein. We define a binding site for a given ligand as the geometric center $\Rz \in \mathbb{R}^3$ of all protein residues within a 4\r{A} radius of any ligand atom.

An overview of all notation used in this work is provided in Appendix~\ref{sec:notation}.

\subsection{Binding Site Prediction Using VN-EGNN}

When experimental binding site annotations are unavailable, accurately identifying binding pockets becomes a critical step in \ac{sbdd}.
\citet{sestak2024vnegnn} proposed an approach based on an \ac{vnegnn} to address this task. In this framework, the protein is represented as a geometric graph (as described above), augmented with a small set of virtual nodes.
Each virtual node is initialized with a coordinate on a sphere surrounding the protein and a feature vector equal to the mean of all protein residue embeddings. Virtual nodes are connected to every protein residue, enabling the network to integrate both local and global structural information.
VN-EGNN employs a three-step heterogeneous message-passing scheme between protein residue nodes $\mathcal{R}$ and virtual nodes $\mathcal{B}$, detailed in Appendix~\ref{sec:vnegnn_message_passing}. The model is trained with a combination of three objective functions (see Appendix~\ref{sec:vnegnn_loss_functions}) to predict the 3D coordinates of potential binding pockets, denoted by the final virtual node positions $\Rz'_1, \ldots, \Rz'_N \in \mathbb{R}^3$, where $N$ is the number of virtual nodes.
In addition to predicting binding site centers, the model outputs pocket-level feature representations $\Bb'_1, \ldots, \Bb'_N \in \mathbb{R}^E$ from the final layer. These embeddings are used to assign confidence scores to predicted pockets and can facilitate downstream tasks such as pocket ranking or contrastive learning.

\subsection{Virtual Screening Using Contrastive Learning}

Contrastive learning provides a general framework for modeling protein–ligand interactions by embedding proteins and ligands into a shared latent space, in which interaction likelihood is assessed via embedding similarity \citep{singh2023contrastive, gao2023drugclip, han2024hashing, wang2024enhancing, mcnutt2024sprint, gilsorribes2025tensordti}.
Most contrastive models in this setting follow a common architectural pattern composed of three core components:
\begin{itemize}[nosep]
    \item a \emph{molecule encoder}, which projects ligand representations into the shared latent space,
    \item a \emph{protein and pocket encoder}, which maps sequence- or structure-based representations of the target protein or binding site into the same space, and
    \item a \emph{contrastive loss function}, that pulls interacting protein–ligand pairs together in the embedding space while pushing non-interacting pairs apart.
\end{itemize}
A defining property of this framework is its separation of representation learning from interaction scoring: once embeddings are computed, virtual screening reduces to efficient similarity evaluations between protein and ligand representations, enabling scalable large-scale screening without explicit docking.

\section{Contrastive Geometric Learning for Unified Drug Design (ConGLUDe)}

\begin{figure*}[t]
  \centering
  \includegraphics[width=0.99\textwidth]{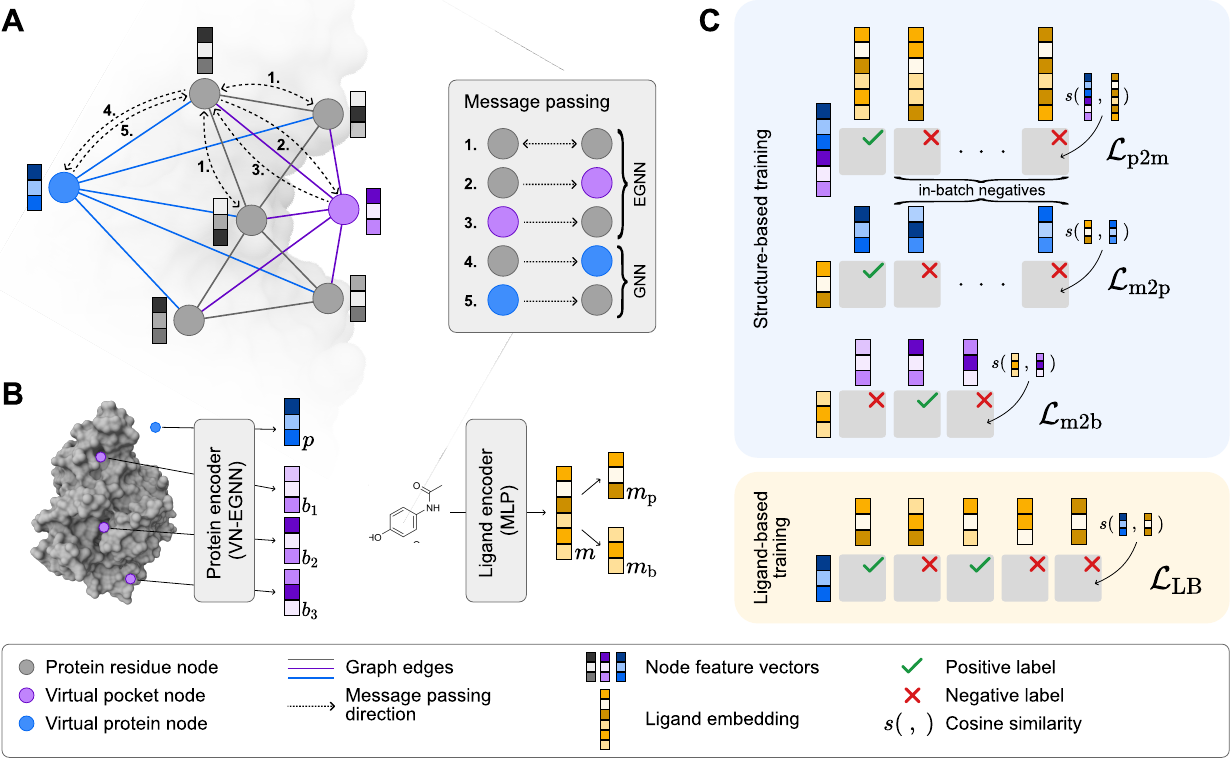}
  \caption{ConGLUDe architecture and training procedure. \textbf{A}: Message-passing scheme of ConGLUDe's protein encoder based on VN-EGNN: 1. message exchange between residue nodes, 2. residue nodes to virtual pocket nodes, 3. pocket nodes
  to residue nodes, 4. residue nodes to virtual protein node, 5. virtual protein node to residue nodes. 
  \textbf{B}: The protein encoder supplies a 
  representation of the whole protein $\Bp$, and
  of each detected pocket $\Bb_k$. The ligand encoder 
  encodes each small molecule into a protein matching representation $\Bm_{\mathrm{p}}$
  and a pocket-matching representation $\Bm_{\mathrm{b}}$. 
  \textbf{C}: Contrastive loss functions used in our approach. Structure-based losses include $\mathcal L_{\mathrm{p2m}}$: InfoNCE between a concatenated protein-pocket representation and all ligand representations from the batch, $\mathcal L_{\mathrm{m2p}}$: InfoNCE between a ligand and all protein representations in the batch, and $\mathcal L_{\mathrm{m2b}}$:
  InfoNCE between a ligand and all pocket representations from the corresponding protein. The NCE loss between a protein and annotated ligand representations ($\mathcal L_{\mathrm{LB}}$) is used on ligand-based data.}
  \label{fig:architecture}
\end{figure*}

ConGLUDe is a contrastive model that employs a geometric \emph{protein encoder} based on a modified VN-EGNN~\citep{sestak2024vnegnn} architecture, which predicts candidate binding site locations $\hat \Rz_1, \ldots, \hat \Rz_K \in \mathbb{R}^3$ together with corresponding representations $\Bb_1, \ldots, \Bb_K \in \mathbb{R}^D$ as well as a global protein embedding $\Bp \in \mathbb{R}^D$. A complementary \emph{molecule encoder} maps ligands into representations $\Bm \in \mathbb{R}^{2D}$, aligned with the concatenated protein/pocket embeddings $[\Bp,\Bb_i]$.

ConGLUDe integrates structure- and ligand-based learning by alternating between (i) structure-based batches, where it learns to detect and characterize binding sites and pair them with their ligands, and (ii) ligand-based batches, where it leverages large-scale bioactivity measurements. Figure~\ref{fig:architecture} provides an overview of the architecture and training procedure.

\subsection{ConGLUDe Architecture}\label{sec:architecture}

\subsubsection{Protein and Binding Pocket Encoders}

We extend the original VN-EGNN formulation by introducing an additional non-geometric virtual node $\mathcal{P}$, which aggregates information from the entire protein but has no spatial coordinates. In addition to the three geometric message-passing steps of VN-EGNN (Appendix~\ref{sec:vnegnn_message_passing}), we add two non-geometric steps from residue nodes $\mathcal{R}$ to the protein node ($\mathcal{R}\to\mathcal{P}$) and vice versa. ($\mathcal{P}\to\mathcal{R}$):

\begin{minipage}[t]{\linewidth}
    \raggedright\textbf{Message passing step 4 ($\mathcal{R} \to \mathcal{P}$)}:
    \begin{align}
        &\Bmu_{j}^{(\mathcal{RP})} = \Bph_{e^{(\mathcal{RP})}}(\Bp, \Bh_j) \label{eq:message4} \\ 
        &\Bmu^{(\mathcal{RP})} = \frac{1}{S} \sum_{j=1}^S \Bmu_{j}^{(\mathcal{RP})} \label{eq:sum4} \\  
        &\Bp = \Bp + \Bph_{h^{(\mathcal{RP})}}\left(\Bp, \Bmu^{(\mathcal{RP})}\right) \label{eq:update4} 
    \end{align}    
\end{minipage}

\begin{minipage}[t]{\linewidth}
    \raggedright\textbf{Message passing step 5 ($\mathcal{P} \to \mathcal{R}$}):
    \begin{align}
        &\Bmu_{i}^{(\mathcal{PR})} = \Bph_{e^{(\mathcal{PR})}}(\Bh_i, \Bp) \label{eq:message5} \\  
        &\Bh_i = \Bh_i + \Bph_{h^{(\mathcal{PR})}}\left(\Bh_i, \Bmu_{i}^{(\mathcal{PR})} \right) \qquad \label{eq:update5} 
    \end{align}
\end{minipage}

Here, $\Bmu_{j}^{(\mathcal{RP})}$ denotes the messages sent from residue node $j$ to the protein node, while $\Bmu_{i}^{(\mathcal{PR})}$ denotes the reverse direction. The functions $\Bph_{e^{(\mathcal{RP})}}$, $\Bph_{h^{(\mathcal{RP})}}$, $\Bph_{e^{(\mathcal{PR})}}$ and $\Bph_{h^{(\mathcal{PR})}}$ are layer-specific \acp{mlp} of the \ac{gnn}. Our model uses 5 layers of \ac{vnegnn}, but we omit the layer index in Eq.~\ref{eq:message1}–\ref{eq:update3} and Eq.\ref{eq:message4}–\ref{eq:update5} for clarity.
Applying the structure encoder to a protein graph $\mathcal{G}$ yields
\begin{equation}
\rX', \BH', \rZ', \BB', \Bp' = \mathrm{VNEGNN}(\mathcal{G}),
\end{equation}
where $\rX'=(\Rx'_1,\ldots,\Rx'_S) \in \mathbb{R}^{S \times 3}$ and $\BH'=(\Bh'_1,\ldots,\Bh'_S) \in \mathbb{R}^{S \times E}$ are the residue coordinates and features, $\rZ'=(\Rz'_1,\ldots,\Rz'_N) \in \mathbb{R}^{N \times 3}$ and $\BB'=(\Bb'_1,\ldots,\Bb'_N) \in \mathbb{R}^{N \times E}$ are the coordinates and feature representations of the virtual nodes representing binding pockets, and $\Bp'$ is the global protein embedding from the protein virtual node.

To rank binding pocket predictions by confidence, we follow \citet{sestak2024vnegnn} and apply a two-layer \ac{mlp} with scalar outputs to the pocket representations: $\Bc' = \mathrm{MLP}(\BB'), \Bc' \in \mathbb{R}^{N}$.
Since multiple virtual nodes may converge to the same binding pocket, we cluster them based on their spatial coordinates using DBSCAN~\citep{ester1996dbscan}. For each cluster, we then compute the mean of the coordinates, feature vectors, and confidence values, yielding $\hat \rZ \in \mathbb{R}^{K \times 3}$, $\hat \BB \in \mathbb{R}^{K \times E}$, and $\hat \Bc \in \mathbb{R}^{K}$ with $K < N$.
Finally, pocket- and protein-level representations are projected into the contrastive embedding space of dimension $D$ via linear transformations:
$\BB = \mathrm{Linear}(\hat \BB), \, \Bp = \mathrm{Linear}(\Bp').$

\subsubsection{Ligand Encoder}

For the ligand encoder, we adopt a simple yet effective design motivated by prior work, which has shown that molecular fingerprints combined with \acp{mlp} often outperform more complex architectures such as \acp{gnn} for encoding small molecules~\citep{unterthiner2014deep, siemers2022minimal, luukkonen2023largescale,seidl2023clamp,praski2025benchmarking}. 
Specifically, a 2-layer \ac{mlp} projects each molecule into a joint 
$2D$-dimensional embedding $\Bm \in \mathbb{R}^{2D}$, which is then split into a protein-matching representation $\Bm_{\mathrm{p}} \in \mathbb{R}^{D}$ and a pocket-matching representation $\Bm_{\mathrm{b}} \in \mathbb{R}^{D}$:
\begin{equation}
    \Bm = [\Bm_{\mathrm{p}}, \Bm_{\mathrm{b}}] = \mathrm{MLP}(\mathcal{M}).
\end{equation}
This lightweight architecture enables simultaneous encoding of large batches of ligands, making it well-suited for high-throughput virtual screening across extensive compound libraries.

\subsubsection{Inference Modes}

ConGLUDe supports multiple inference modes. In classical \emph{virtual screening}, predictions are made by comparing the protein representation with the protein-specific component of the ligand embedding, $s\left(\Bp, \Bm_{\mathrm{p}}\right)$, where $s(.,.)$ denotes the cosine similarity and higher similarity indicates a higher likelihood of binding. This formulation also applies to \emph{target fishing}, where multiple proteins are ranked for a given small molecule.
For \emph{binding site identification}, the VN-EGNN–based encoder directly outputs candidate pocket centers with confidence values. Predicted pockets can be ranked either ligand-independently by these confidence scores or in a ligand-conditioned manner by their similarity to the pocket-specific component of the ligand embedding, $s\left(\Bb_l, \Bm_{\mathrm{b}}\right)$.

\subsection{ConGLUDe Training}

\subsubsection{Data}

The ConGLUDe architecture enables training on a 
ligand-protein pairs from 
both structure-based and ligand-based databases. 
For each 
task, structure-based training data, a subset of 
PDBbind \citep{wang2005pdbbind}, are derived from 
the respective baseline methods. As ligand-based data 
we use the MERGED dataset
\citep{golts2024large, mcnutt2024sprint}, which combines PubChem 
\citep{kim2025pubchem}, BindingDB 
\citep{gilson2015bindingdb}, and ChEMBL 
\citep{gaulton2011chembl}, and remove all proteins with 
>90\% sequence identity to any test set protein. For 
details on all datasets, see Appendix~\ref{app:sec:datasets}.

\subsubsection{Training Objective}

ConGLUDe is trained by minimizing 
\begin{align}
\mathcal L &= \mathcal{L}_{\mathrm{SB}} + \mathcal{L}_{\mathrm{LB}} ,
\end{align}
where $\mathcal{L}_{\mathrm{SB}}$ and $\mathcal{L}_{\mathrm{LB}}$ are the structure- and ligand-based loss functions, further detailed below. At each training step, a batch 
of either structure-based or ligand-based data is sampled at random, and the optimization objective is applied accordingly.

\textbf{Training on structure-based data.}
For structure-based data, annotated protein binding sites provide 
supervision for binding site prediction. 
In this setting, the loss decomposes into a geometric term and a contrastive term:
\begin{align}
    \mathcal L_{\mathrm{SB}} &= \mathcal{L}_{\mathrm{geometric}} + \mathcal{L}_{\mathrm{contrastive}} .
\end{align}
The geometric component, $\mathcal{L}_{\mathrm{geometric}} .$ 
is equivalent to the objective function of \ac{vnegnn} (see Appendix~\ref{sec:vnegnn_loss_functions}).

Beyond the geometric objective, we leverage contrastive learning 
to align the representations of ligands with their corresponding proteins 
and predicted binding pockets. For a given protein-ligand complex, 
the ligand embedding $\Bm^{j}$ is encouraged to be close in 
representation space to the concatenated protein and pocket embeddings 
$[\Bp^{j}, \Bb_l^{j}]$, where $\Bb_l^{j}$ 
is the predicted pocket closest to the ligand’s true binding site center $\Rz$:
$l=\text{argmin}_{k=1,\ldots,K}(||\Rz-\hat{\Rz}_k||)$.

This alignment is implemented using a three-way InfoNCE loss (Eq.~\ref{eq:infonce}), similar to CLIP \citep{radford2021learning}, aligning protein, pocket, and ligand embeddings along complementary axes.
Along the first axis -- “protein+pocket to molecule” -- the concatenated protein and pocket representation, $[\Bp^{j},\Bb_l^{j}]$, acts as the anchor, and the model is trained to associate it with its true ligand $\Bm^{j}$ while treating other ligands in the batch as negatives: 
\begin{equation}
    \mathcal{L}^{j}_{\mathrm{p2m}} = \mathrm{InfoNCE}([\Bp^{j},\Bb_l^{j}], \Bm^{j}, \{\Bm^{i}\}_{i=1}^{J}; \tau_{\mathrm{p2m}}).
\end{equation}
The second axis aligns the protein-matching component of the ligand embedding, $\Bm_{\mathrm{p}}^{j}$, with the global protein representation $\Bp^{j}$, while contrasting it against other proteins in the mini-batch (“molecule to protein”):
\begin{equation}
    \mathcal{L}^{j}_{\mathrm{m2p}} = \mathrm{InfoNCE}(\Bm_{\mathrm{p}}^{j}, \Bp^{j}, \{\Bp^{i}\}_{i=1}^{J}; \tau_{\mathrm{m2p}}).
\end{equation}
Along the third axis -- "molecule to binding site" -- the pocket-matching component of the ligand embedding, $\Bm_{\mathrm{b}}^{j}$, is aligned with the closest predicted binding pocket $\Bb_l^{j}$, contrasting it against the remaining predicted pockets on the same protein:
\begin{equation}
    \mathcal{L}^{j}_{\mathrm{m2b}} = \mathrm{InfoNCE}(\Bm_{\mathrm{b}}^{j}, \Bb_l^{j}, \{\Bb_k^{j}\}_{k=1}^{K}; \tau_{\mathrm{m2b}}).
\end{equation}
The temperature parameters are chosen as the inverse square root of the 
corresponding contrastive space dimension, 
i.e. $\tau_{\mathrm{p2m}} = \frac{1}{\sqrt{2D}}$ 
and $\tau_{\mathrm{m2p}} = \tau_{\mathrm{m2b}} = \frac{1}{\sqrt{D}}$.
The total contrastive loss is obtained by summing up the three loss axes and averaging over all samples in the batch:
\begin{equation}
    \mathcal{L}_{\mathrm{contrastive}} = \frac{1}{J} \sum_{j=1}^J \left (\mathcal{L}^{j}_{\mathrm{p2m}} +\mathcal{L}^{j}_{\mathrm{m2p}} + \mathcal{L}^{j}_{\mathrm{m2b}} \right).
\end{equation}
Alternative contrastive formulations, such as the CLOOB loss \citep{furst2022cloob, sanchez2023cloome}, could also be used.

\textbf{Training on ligand-based data.}
When training on ligand-based datasets, we leverage large collections 
of annotated active and inactive compounds for a given protein target.
Since no structural information about the binding pocket is available in this setting, the VN-EGNN module cannot be meaningfully optimized and is therefore kept frozen during training, with only the linear layer on the protein node and the ligand encoder being trained.
For each batch, active and inactive compounds are sampled 
at a ratio of 1:3, and the model is trained with a \emph{sigmoid contrastive loss} \citep{gutmann2010noise,seidl2023clamp,zhai2023sigmoid}, 
which uses the cosine similarity between the 
whole-protein representation $\Bp$ and the corresponding 
part of the small molecule embeddings $\Bm_{\mathrm{p}}$, and 
the activity labels $\By_m \in \{0,1\}$:
\begin{align}
    \mathcal{L}_{\mathrm{LB}}&( \Bp,\{\Bm_{\mathrm{p}}^{m}\}_{m=1}^{M}, \{y_m\}_{m=1}^{M})   \\
    &= - \frac{1}{M} \sum_{m=1}^M \Big[
        y_m \log(q_m) + (1 - y_m) \log(1 - q_m)
    \Big], \nonumber
\end{align}
where $q_m = \sigma\bigl(s(\Bp, \Bm_{\mathrm{p}}^m)\bigr)$, with the sigmoid function $\sigma$.

\section{Experiments and Results}

\newcommand{\NA}{\multicolumn{1}{c}{--}}

\begin{table*}[t]
    \centering
    \setlength{\tabcolsep}{8pt}
    \begin{threeparttable}
        \caption{Zero-shot virtual screening performance on the LIT-PCBA datasets.
        For ConGLUDe, we report the median and mean absolute deviation (MAD) over three training re-runs. The best value per column as well as all other values within one MAD are marked in bold.}
        \label{tab:lit_pcba_main}
        \footnotesize
        \begin{tabular}{
            l
            l
            S[table-format=2.2]
            S[table-format=2.2]
            S[table-format=2.2]
            S[table-format=2.2]
            S[table-format=2.2]
        }
        \toprule
        & & {AUROC $\uparrow$} & {BEDROC $\uparrow$} & {EF $0.5 \%\uparrow$} & {EF $1 \%\uparrow$} & {EF $5 \%\uparrow$} \\
        \midrule
        $\dagger$ & DrugCLIP \citep{gao2023drugclip} & 57.17 & 6.23 & 8.56 & 5.51 & 2.27 \\
        $\dagger$ & DrugHash \citep{han2024hashing}  & 54.58 & 7.14 & 9.65 & 6.14 & 2.42 \\
    	$\dagger$ & S$^2$Drug \citep{he2025s2drug} & 58.23 & 8.69 & 11.44 & 7.38 & 2.97 \\
        $\dagger$ & LigUnity \citep{feng2025hierarchical}  & 59.85 & {\bfseries11.33} & \NA & 6.47 & \NA \\
        $\dagger$ & HypSeek \citep{wang2025hypseek} & 62.10 & {\bfseries 11.96} & \NA & 6.81 & \NA \\
        \midrule
        & DrugCLIP\textsubscript{P2Rank}\tnote{a} & 51.23 & 2.65 & 2.44 & 1.80 & 1.36 \\
        & DrugCLIP\textsubscript{VN-EGNN}\tnote{a} & 56.70 & 4.15 & 2.51 & 3.17 & 2.04 \\
        & SPRINT \citep{mcnutt2024sprint} & {\bfseries 73.40} & {\bfseries 12.30} & {\bfseries 15.90} & {\bfseries 10.78} & {\bfseries \;\;5.29} \\
        \midrule
        & ConGLUDe 
            & 64.06\tnote{$\pm 3.25$} 
            & {\bfseries 12.24}\tnote{$\pm 2.06$} 
            & {\bfseries 15.87}\tnote{$\pm 2.06$}
            & {\bfseries 11.03}\tnote{$\pm 1.81$}
            & 4.68\tnote{$\pm 0.30$} \\
        \bottomrule
        \end{tabular}
        \begin{tablenotes}
            \footnotesize
            \item[$\dagger$] Known binding pocket given as input.
            \item[a] Evaluated in this work.
        \end{tablenotes}        
    \end{threeparttable}
\end{table*}

We evaluate ConGLUDe on four drug discovery tasks: virtual screening (Section~\ref{sec:virtual_screening}), target fishing (Section~\ref{sec:target_fishing}), binding pocket prediction (Section~\ref{sec:binding_site_prediction}), and ligand-conditioned pocket selection (Section~\ref{sec:pocket_selection}). For each task, we train a separate model with task-specific data splits to ensure that the corresponding test data are excluded from training. We compare against established benchmarks for virtual screening and binding site prediction, and propose new evaluation protocols for target fishing and pocket selection tasks, which lack standardized benchmarks, by assessing the relevant baseline methods ourselves.

Details on the data, training procedure, and task-specific evaluation metrics are provided in Sections~\ref{app:sec:datasets}, \ref{app:sec:train_details}, and \ref{app:sec:metrics}, respectively. Ablation studies and extended results are reported in Sections~\ref{app:sec:ablation} and \ref{app:sec:extended_results}.

\subsection{Virtual Screening}\label{sec:virtual_screening}

We compare ConGLUDe against several contrastive learning–based approaches. The baseline methods fall into two categories. First, models with Uni-Mol-based encoders \citep{zhou2023unimol,gao2023drugclip,han2024hashing, he2025s2drug, feng2025hierarchical, wang2025hypseek} rely on explicit pocket representations and require the binding pocket as input. Second, pocket-agnostic methods do not assume prior pocket information. This group includes SPRINT \citep{mcnutt2024sprint}, which encodes the entire protein, as well as two hybrid approaches implemented by us that combine DrugCLIP with a pocket predictor, either P2Rank \citep{krivak2018p2rank} or VN-EGNN \citep{sestak2024vnegnn}. For these hybrids, the top-ranked pocket predicted by the respective method is selected and encoded using DrugCLIP.
We report results on LIT-PCBA \citep{trannguyen2020litpcba} using several metrics in Table \ref{tab:lit_pcba_main}. For completeness, Section \ref{app:sec:vs} includes comparisons to classical docking and deep learning based methods, and also reports results on DUD-E \citep{mysinger2012directory}, a widely used but flawed benchmark \citep{chen2019hidden}. On LIT-PCBA, we substantially outperform methods that use binding site information as input and match SPRINT, which also encodes the full protein. However, we markedly outperform SPRINT on the more structure-centric DUD-E benchmark and at target fishing.

\subsection{Target Fishing}\label{sec:target_fishing}

Target fishing refers to the task of identifying protein targets for a given small molecule ligand. We evaluate ConGLUDe on a target fishing dataset from \citet{reinecke2024chemical}, which contains experimentally determined targets for approximately 1,000 ligands. These targets were identified using Kinobeads chemical proteomics, an experimental technology that differs substantially from the data sources used to train ConGLUDe. As a result, this dataset represents a challenging out-of-domain setting, which we address in a zero-shot manner.

\begin{table}[h]
    \centering
    \caption{
        Zero-shot performance on the target fishing task measured by mean AUROC, $\Delta$AUPRC, and EF at 1\% averaged across small molecules. The best value per column is marked in bold. ConGLUDe significantly outperforms the second-best method (Wilcoxon test, $p \approx 10^{-24}$).}
    \label{tab:target_fishing}
    \footnotesize
    \begin{threeparttable}
        \begin{tabular}{lccc}
            \toprule
             & AUROC $\uparrow$ & $\Delta$AUPRC $\uparrow$ & EF $1 \%\uparrow$ \\
            \midrule
            DrugCLIP\textsubscript{P2Rank} & $51.2$ & $0.6$ & $1.4$ \\
            DrugCLIP\textsubscript{VN-EGNN} & $54.2$ & $0.4$ & $0.8$ \\
            SPRINT & $42.5$ & $0.3$ & $0.8$  \\
            DiffDock & $58.9$ & $2.2$ & $5.3$ \\
            \midrule
            ConGLUDe & $\textbf{65.6}$ & $\textbf{5.1}$ & $\textbf{9.9}$ \\
            \bottomrule
        \end{tabular}
    \end{threeparttable}
\end{table}

In Table~\ref{tab:target_fishing}, we compare ConGLUDe to DrugCLIP \citep{gao2023drugclip}, SPRINT \citep{mcnutt2024sprint}, and DiffDock \citep{corso2023diffdock}. Since DrugCLIP requires pocket structures, which are unavailable for this dataset, we generate candidate pockets using P2Rank and VN-EGNN. ConGLUDe consistently outperforms all baselines, and the difference in per-ligand AUCs between ConGLUDe and DiffDock is highly significant (Wilcoxon test, $p \approx 10^{-24}$). All other methods perform close to random, highlighting the challenge of the task. DiffDock, while the strongest baseline, requires multiple GPU-days for this evaluation, making it impractical for large-scale virtual screening or target fishing.

\subsection{Binding Site Prediction}\label{sec:binding_site_prediction}

As shown in Table \ref{tab:pocket_prediction}, we retain the performance of \ac{vnegnn} on binding site prediction which highlights the robustness of the VN-EGNN encoder. The architectural modifications and adaptations to support additional tasks do not hinder its ability to perform well on pocket prediction. Full results of compared methods from \citet{sestak2024vnegnn} are in Appendix Table \ref{tab:app_pp_results}.

\begin{table}[h]
    \centering
    \begin{threeparttable}
    \caption{Performance at binding site identification measured by the top-1 DCC success rate at a 4\r{A} threshold ($\uparrow$) on the COACH420, HOLO4K, and PDBbind datasets. Best value marked bold.}
    \label{tab:pocket_prediction}
    \footnotesize
    \begin{tabular}{lccc}
        \toprule
        & COACH420 & HOLO4K & PDBbind2020 \\
        \midrule
        VN-EGNN   & \textbf{0.605} & \textbf{0.532} & 0.669  \\
        ConGLUDe  & 0.602 & 0.525 & \textbf{0.689} \\
        \bottomrule
    \end{tabular}
    \end{threeparttable}
\end{table}

\subsection{Ligand-Conditioned Pocket Selection}
\label{sec:pocket_selection}

Finally, we evaluate \emph{ligand-conditioned pocket selection}, where candidate pockets are ranked by their likelihood of binding a given ligand, in contrast to unconditioned predictors that ignore ligand information. 
We compare against two unconditioned baselines, P2Rank \citep{krivak2018p2rank} and VN-EGNN \citep{sestak2024vnegnn}, as well as two-step approaches that pair DrugCLIP \citep{gao2023drugclip} with an unconditioned pocket predictor, ranking DrugCLIP-encoded pockets by their similarity with corresponding ligand embeddings. 
We further include DiffDock \citep{corso2023diffdock} and AutoDock Vina \citep{trott2010vina}, both evaluated as blind docking methods that implicitly select binding pockets during docking, as well as AutoDock Vina paired with VN-EGNN-predicted pockets, analogous to the DrugCLIP two-step approach. Unlike docking, which explicitly simulates each ligand–pocket pair, ConGLUDe and DrugCLIP embed ligands and pockets independently and score them via a dot product, resulting in a substantial speed advantage.

We evaluate on the PDBbind time split \citep{stark2022equibind}, PoseBusters \citep{buttenschoen2024posebusters}, and the Allosteric Site Database (ASD) \citep{liu2020unraveling}, reporting the top-1 DCC success rate at 4\r{A}.
ConGLUDe achieves top performance on all three datasets (Table \ref{tab:pocket_selection}).

\begin{table}[h]
    \centering
    \caption{Performance of ligand-conditioned pocket selection measured by the top-1 DCC success rate at a 4\r{A} threshold ($\uparrow$). Values in parentheses indicate 95\% confidence intervals. The best-performing method is highlighted in bold. DC = DrugCLIP.}
    \label{tab:pocket_selection}
    \footnotesize
    \begin{threeparttable}
        \begin{tabular}{lccc}
            \toprule
             & PDBbind Time & PoseBusters & ASD \\
            \midrule
            P2Rank & 0.49$^{(.43,.56)}$ & 0.20$^{(.17,.24)}$ & 0.25$^{(.23,.27)}$ \\
            VN-EGNN  & 0.51$^{(.45,.57)}$ & \textbf{0.27}$^{(.23,.31)}$ & 0.28$^{(.26,.30)}$ \\
            \midrule
            DiffDock & 0.35$^{(.30,.42)}$ & 0.15$^{(.12,.18)}$ & \textbf{0.35}$^{(.33,.37)}$\\
            Vina & 0.36$^{(.32,.41)}$ & 0.16$^{(.14,.18)}$ & 0.20$^{(.18,.21)}$ \\
            \midrule
            DC\textsubscript{P2Rank} & 0.08$^{(.05,.12)}$ & 0.04$^{(.02,.06)}$ & 0.12$^{(.11,.14)}$ \\
            DC\textsubscript{VN-EGNN} & 0.39$^{(.33,.46)}$ & 0.19$^{(.16,.23)}$ & 0.17$^{(.16,.19)}$\\
            Vina\textsubscript{VN-EGNN} & 0.52$^{(.47,.57)}$ & 0.25$^{(.22,.27)}$ & 0.31$^{(.29,.35)}$ \\
            \midrule
            ConGLUDe & \textbf{0.54}$^{(.47,.60)}$ & \textbf{0.27}$^{(.24,.31)}$ & \textbf{0.35}$^{(.33,.37)}$\\
            \bottomrule
        \end{tabular}
    \end{threeparttable}
\end{table}

Notably, on the PDBbind and PoseBusters datasets, ConGLUDe shows limited improvement over the inherent VN-EGNN ranking. We attribute this to the predominance of orthosteric ligands in PDB-derived data, which unconditional pocket predictors already rank highly. In contrast, on ASD, which contains rarer allosteric binding sites where ligand specificity is crucial, ConGLUDe significantly enhances ligand-conditioned pocket selection compared to the unconditioned baseline. By comparison, DrugCLIP, which was not trained to distinguish among multiple pockets on a protein, fails to yield a meaningful ranking.

\subsection{Inference Speed}
\begin{figure}[h]
    \centering
    \includegraphics[width=\linewidth]{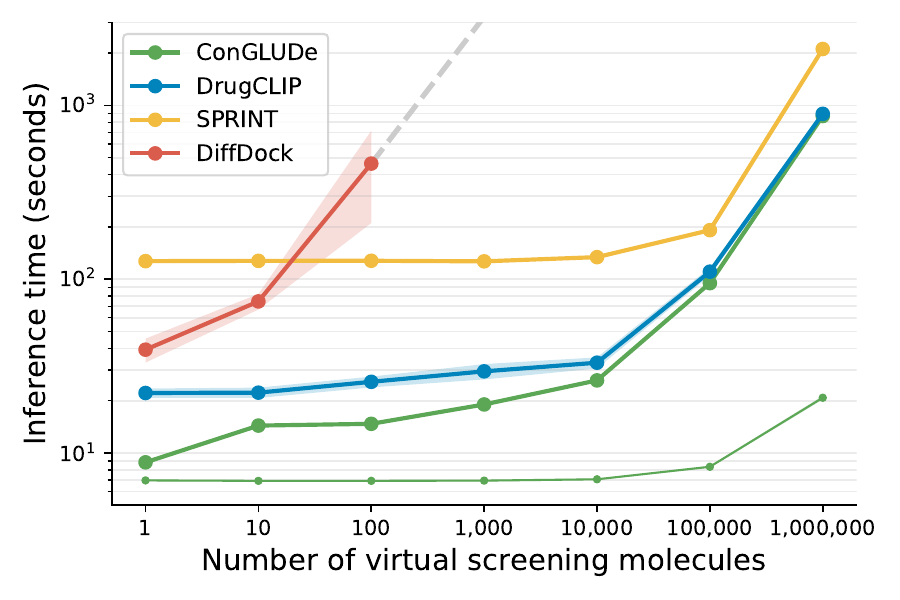}
    \caption{Average virtual screening time per protein target versus the number of screened molecules, averaged over 100 targets with error bars indicating standard deviations. For ConGLUDe, the thin line uses precomputed ligand fingerprints and descriptors, while the thick line includes preprocessing. SPRINT and DiffDock report end-to-end time including preprocessing; DrugCLIP uses precomputed pocket and ligand coordinates.}
    \label{fig:timing}
\end{figure}

To demonstrate the efficiency of our approach, we performed a timing analysis in the virtual screening setting against three baselines: the contrastive methods SPRINT and DrugCLIP, and DiffDock as a representative docking method. Average virtual screening times for varying numbers of ligands on a single protein target are shown in Figure \ref{fig:timing}. 
While contrastive learning methods can efficiently screen millions of compounds, docking becomes expensive even for several thousand ligands. Moreover, adding more protein targets does not substantially increase computation time for contrastive methods, since precomputed ligand embeddings can be reused, whereas DiffDock requires a full rerun for each target. 

For ConGLUDe and SPRINT, the main computational bottleneck is the calculation of RDKit representations of ligands, as illustrated by the difference between the thin line (no preprocessing) and the thick line (with preprocessing) for ConGLUDe. Notably, even with preprocessing, ConGLUDe matches the inference speed of DrugCLIP, which uses already computed pocket and molecule conformer coordinates.

\section{Conclusion}\label{sec:limitations_and_discussion}

We present ConGLUDe, a contrastive learning-based method for drug discovery, which enables the unified training on structure- and ligand-based data by integrating a geometric binding site prediction module directly within the protein encoder. With this approach, ConGLUDe achieves state-of-the-art performance across a variety of both traditionally structure- and ligand-based tasks with high computational efficiency, demonstrating its broad applicability.

\textbf{Limitations.} While ConGLUDe performs well for proteins with experimentally resolved 3D structures as found in the PDB, its behavior on predicted structures or proteins highly divergent from known templates remains uncertain. Although ConGLUDe can identify ligand-specific pockets substantially faster than traditional docking or co-folding approaches, it does not generate docked ligand conformations. For applications that require explicit docking, ConGLUDe-predicted pockets can nevertheless serve as an efficient starting point for standard docking workflows. Moreover, its extension to ligand-based data currently assumes bioassays with a uniquely associated protein target and does not directly support phenotypic or target-agnostic assays.

\textbf{Outlook.} Despite its already strong performance across diverse tasks, ConGLUDe could be extended in several promising directions. Integrating generative models could enable joint prediction and design of ligands for specific pockets. The framework could also be extended to the prediction of additional properties such as binding affinity or ADMET profiles. More generally, our approach suggests a path toward general-purpose foundation models for drug discovery that jointly learn from structural protein-ligand complexes and large-scale bioactivity data.

\section*{Software and Data}
The code for ConGLUDe is available at \url{https://github.com/ml-jku/conglude}.

\section*{Acknowledgements}
The ELLIS Unit Linz, the LIT AI Lab, the Institute for Machine Learning, are supported by the Federal State Upper Austria. We thank the projects FWF AIRI FG 9-N (10.55776/FG9), AI4GreenHeatingGrids (FFG- 899943), Stars4Waters (HORIZON-CL6-2021-CLIMATE-01-01), FWF Bilateral Artificial Intelligence (10.55776/COE12). We thank NXAI GmbH, Audi AG, Merck Healthcare KGaA, T\"{U}V Holding GmbH, Software Competence Center Hagenberg GmbH, dSPACE GmbH, TRUMPF SE + Co. KG.

\section*{Impact Statement}

This paper presents work aimed at advancing machine learning methods for drug discovery. While our methods have potential applications in biomedical research, we do not identify any societal consequences that require specific highlighting here.

\clearpage
\bibliography{references}
\bibliographystyle{icml2026}

\appendix
\onecolumn

\counterwithin{figure}{section}
\counterwithin{table}{section}
\counterwithin{equation}{section}
\renewcommand\thefigure{\thesection\arabic{figure}}
\renewcommand\thetable{\thesection\arabic{table}}
\renewcommand\theequation{\thesection.\arabic{equation}}

\section{Notation}
\label{sec:notation}

The following table summarizes all notation used throughout this paper.

\resizebox{\textwidth}{!}{%
\small
\begin{tabular}{l c l}
\toprule
Definition & Symbol & Type  \\ 

\midrule
\multicolumn{3}{l}{\textbf{Scalars}} \\ 
batch size & $J$ & $\mathbb{N}$ \\
contrastive space dimension & $D$ & $\mathbb{N}$\\
VN-EGNN output dimension & $E$ & $\mathbb{N}$ \\
number of binding site VNs & $N$ & $\mathbb{N}$ \\
number of predicted binding sites after clustering & $K$ & $\mathbb{N}$ \\
number of labeled small molecules for a given protein & $M$ & $\mathbb{N}$ \\
number of protein residues & $S$ & $\mathbb{N}$ \\

\midrule
\multicolumn{3}{l}{\textbf{Representations}} \\ 
protein residue representation & $\Bh_s'$ & $\mathbb{R}^E$ \\
pocket representation before clustering & $\Bb_n'$ & $\mathbb{R}^E$ \\
protein representation before projection & $\Bp'$ & $\mathbb{R}^E$ \\
pocket representation before projection & $\hat \Bb_k$ & $\mathbb{R}^E$ \\
final protein representation & $\Bp$ & $\mathbb{R}^D$ \\
final pocket representation & $\Bb_k$ & $\mathbb{R}^D$ \\
small molecule representation & $\Bm_m = [\Bm_{\mathrm{p}m}, \Bm_{\mathrm{b}m}]$ &  $\mathbb{R}^{2D}$ \\

\midrule
\multicolumn{3}{l}{\textbf{Coordinates}} \\ 
protein residue position & $\Rx_s'$ & $\mathbb{R}^3$ \\
pocket node position before clustering & $\Rz_n'$ & $\mathbb{R}^3$ \\
predicted binding pocket center/final VN position & $\hat \Rz_k$ & $\mathbb{R}^3$ \\
ground-truth binding pocket center & $\Rz$ & $\mathbb{R}^3$ \\

\midrule
\multicolumn{3}{l}{\textbf{Data quantities}} \\ 
predicted confidence value for $\hat \Rz_k$ & $\hat c_k$ & $\mathbb{R}$ \\
ground-truth confidence value for $\hat \Rz_k$ & $c_k$ & $\{c_0,\left[0.5,1\right]\}$ \\
residue-level binding site label & $z_s$ & $\{0,1\}$ \\
binary activity label for molecule $\Bm_m$ & $y_m$ & $\{0,1\}$ \\

\midrule
\multicolumn{3}{l}{\textbf{Constants}} \\ 
fall-back value for confidence calculation & $c_0$ & 0.001 \\
tolerance radius for confidence calculation & $\gamma$ & 4.0 \\
temperature for $\mathcal{L}_{\mathrm{p2m}}$ & $\tau_{\mathrm{p2m}}$  & $\frac{1}{\sqrt{2D}}$\\
temperature for $\mathcal{L}_{\mathrm{m2p}}$ & $\tau_{\mathrm{m2p}}$ & $\frac{1}{\sqrt{D}}$\\
temperature for $\mathcal{L}_{\mathrm{m2b}}$ & $\tau_{\mathrm{m2b}}$ & $\frac{1}{\sqrt{D}}$\\
\midrule
\multicolumn{3}{l}{\textbf{Functions}} \\ 
cosine similarity & $s(.,.)$ & $\mathbb{R}^D \times \mathbb{R}^D \rightarrow \left[-1,1\right]$  \\
sigmoid function & $\sigma(.)$ & $\mathbb{R} \rightarrow \left[0,1\right]$ \\
\bottomrule
\end{tabular}} \\

The InfoNCE loss used for structure-based training is defined as follows:
\begin{align} 
    \label{eq:infonce}
    \mathrm{InfoNCE}(\Bq^{j},\Bk^{j}, \{\Bk^{i}\}_{i=1}^J; \tau) = - \log \frac{\exp\left(s(\Bq^{j}, \Bk^{j})/\tau \right)}{\sum_{i=1}^J \exp\left(s(\Bq^{j},\Bk^{i})/\tau \right)} .
\end{align}

\section{VN-EGNN Details}
\label{sec:vnegnn}

\subsection{Heterogeneous Message Passing}
\label{sec:vnegnn_message_passing}

Following \citet{sestak2024vnegnn}, we briefly summarize the heterogeneous message passing scheme used in VN-EGNN. Each layer consists of three message passing steps that exchange information between protein residues ($\mathcal{R}$) and virtual binding pocket nodes ($\mathcal{B}$).

The first step corresponds to the standard \ac{egnn} formulation \citep{satorras2021egnn}, where information is exchanged between neighboring protein residues:

\textbf{Message passing step 1 ($\mathcal{R} \rightarrow \mathcal{R}$)}:
    \begin{align}
        &\Bmu_{ij}^{(\mathcal{RR})} = \Bph_{e^{(\mathcal{RR})}}(\Bh_i,\Bh_j,\lVert \Rx_i - \Rx_j \rVert) \label{eq:message1} \\ 
        &\Bmu_i^{(\mathcal{RR})} = \frac{1}{|\mathcal N(i)|} \sum_{j\in \mathcal N(i)} \Bmu_{ij}^{(\mathcal{RR})} \label{eq:sum1}\\  
        &\Rx_i = \Rx_i + \frac{1}{|\mathcal N(i)|} \sum_{j\in \mathcal N(i)} \frac{\Rx_i - \Rx_j}{\lVert \Rx_i - \Rx_j\rVert} \phi_{\Rx^{(\mathcal{RR})}}(\Bmu_{ij}^{(\mathcal{RR})}) \label{eq:x_coord1}  \\
        &\Bh_i = \Bh_i + \Bph_{h^{(\mathcal{RR})}}\left(\Bh_i, \Bmu_i^{(\mathcal{RR})}\right). \label{eq:update1} 
    \end{align}

Here, the coordinates $\Rx_i$ and features $\Bh_i$ of residue nodes are updated based on aggregated messages from their neighbors. The \acp{mlp} $\Bph_{e^{(\mathcal{RR})}}$, $\phi_{\Rx^{(\mathcal{RR})}}$, and $\Bph_{h^{(\mathcal{RR})}}$ are learnable functions specific to each layer. The same applies to all MLPs $\Bph_{.}$ in the subsequent steps.

In the second step, residue nodes transmit information to virtual pocket nodes $\mathcal{B}$, which act as proxies for potential binding sites:

\textbf{Message passing step 2 ($\mathcal{R} \rightarrow \mathcal{B}$)}:
    \begin{align}
        &\Bmu_{ij}^{(\mathcal{RB})} = \Bph_{e^{(\mathcal{RB})}}(\Bb_i, \Bh_j, \lVert \Rz_i - \Rx_j\rVert) \label{eq:message2}
        \\ &\Bmu_i^{(\mathcal{RB})} = \frac{1}{S} \sum_{j=1}^S \Bmu_{ij}^{(\mathcal{RB})} \label{eq:sum2} \\  
        &\Rz_i = \Rz_i + \frac{1}{S} \sum_{j=1}^S \frac{\Rz_i - \Rx_j} {\lVert \Rz_i - \Rx_j \rVert} \phi_{\Rx^{(\mathcal{RB})}}(\Bmu_{ij}^{(\mathcal{RB})})  \label{eq:z_coord}
        \\ &\Bb_i = \Bb_i + \Bph_{h^{(\mathcal{RB})}}\left(\Bb_i, \Bmu_i^{(\mathcal{RB})}\right) \label{eq:update2} 
    \end{align}

Finally, the third step propagates information in the reverse direction, from virtual nodes back to residue nodes:

\textbf{Message passing step 3 ($\mathcal{B} \rightarrow \mathcal{R}$)}:
    \begin{align}
        &\Bmu_{ij}^{(\mathcal{BR})} = \Bph_{e^{(\mathcal{BR})}}(\Bh_i, \Bb_j, \lVert \Rx_i - \Rz_j \rVert) \label{eq:message3} 
        \\ &\Bmu_i^{(\mathcal{BR})} = \frac{1}{N} \sum_{j=1}^N \Bmu_{ij}^{(\mathcal{BR})} \label{eq:sum3} \\ 
            &\Rx_i = \Rx_i + \frac{1}{N} \sum_{j=1}^N \frac{\Rx_i - \Rz_j}{\lVert \Rx_i - \Rz_j \rVert} \phi_{\Rx^{(\mathcal{BR})}}(\Bmu_{ij}^{(\mathcal{BR})})
        \\  &\Bh_i = \Bh_i + \Bph_{h^{(\mathcal{BR})}}\left(\Bh_i, \Bmu_i^{(\mathcal{BR})}\right) \label{eq:update3} 
    \end{align}

\subsection{Objective Functions}
\label{sec:vnegnn_loss_functions}

VN-EGNN \citep{sestak2024vnegnn} is trained using a combination of losses that supervise the prediction of binding site centers, residue-level segmentation, and confidence of the predictions.

To ensure accurate prediction of the binding site center (bsc) location, the squared distance between the true binding site center $\Rz$ and the closest predicted center $\Rz_n'$ among $N$ candidates is minimized:

\begin{align}
    \mathcal{L}_{\mathrm{bsc}}(\{\Rz_1',\ldots,\Rz_N'\},\Rz) = \min_{n \in {1,\ldots,N}} \lVert \Rz - \Rz_n' \rVert^2.
\end{align}

For residue-level binding site segmentation, the network outputs predictions for each residue $s$ through a multilayer perceptron: $\hat z_s = MLP(\Bh_s')$. The segmentation loss is defined as a differentiable Dice loss, which compares the predicted and true residue labels $z_s$:

\begin{align}
    \mathcal{L}_{\mathrm{seg}}(\{\hat z_1,\ldots,\hat z_S\}, \{z_1,\ldots, z_S\}; \epsilon) = 1 - \frac{2 \ \sum_{s=1}^S z_s \ \hat z_s + \epsilon}{\sum_{s=1}^S z_s + \sum_{s=1}^S \hat z_s + \epsilon},
\end{align}

where $\epsilon$ is a small constant to stabilize the division.

Moreover, each predicted center $\Rz_n'$ is assigned a confidence score $c_n'$ which should reflect its proximity to the true center. The target confidence $c_n$ is defined as:

\begin{align}
    c_n = 
        \begin{cases} 
        1 - \frac{1}{2\gamma} \cdot \lVert \Rz - \Rz_n' \rVert & \text{if }\lVert \Rz - \Rz_n' \rVert \leq \gamma, \\
        c_0 & \text{otherwise,}
        \end{cases},
\label{eq:confidence}
\end{align}

and the corresponding confidence loss is the mean squared error between predicted and target confidences:

\begin{align}
    \mathcal{L}_{\text{confidence}}(\{c_1',\ldots,c_N'\}, \{c_1,\ldots, c_N\}) = \frac{1}{N}\sum_{n=1}^N (c_n-c_n')^2.
\end{align}

The total VNEGNN objective combines the three components and is used as the geometric learning objective in ConGLUDe's structure-based training:

\begin{align}
    \mathcal{L}_{\mathrm{geometric}} = \mathcal{L}_{\mathrm{bsc}} + \mathcal{L}_{\mathrm{seg}} + \mathcal{L}_{\mathrm{confidence}} .
\end{align}

\clearpage
\section{Datasets}\label{app:sec:datasets}

\subsection{Structure-Based Training Datasets}

For structure-based training, we utilized subsets of PDBbind v.2020 \citep{wang2005pdbbind}, adopting the dataset partitions established by the baseline methods corresponding to each task. Specifically, for virtual screening and target fishing, we followed the DrugCLIP split \citep{gao2023drugclip}. For binding site prediction, we trained on scPDB, consistent with VN-EGNN \citep{sestak2024vnegnn}, and for ligand-conditioned pocket selection, we employed the time-based split used in DiffDock \citep{corso2023diffdock}.

\subsection{Ligand-Based Training Datasets}

For training on ligand-based data, we employed the MERGED dataset \citep{golts2024large, mcnutt2024sprint}, which integrates data from PubChem \citep{kim2025pubchem}, BindingDB \citep{gilson2015bindingdb}, and ChEMBL \citep{gaulton2011chembl}. We use the combined MERGED training and test splits as the basis for our training set and keep the same validation split as \citet{mcnutt2024sprint}. For each protein in the dataset, we used an available 3D structure from the PDB; if none was available, we used an AlphaFold2 \citep{jumper2021alphafold} structure instead.
To prevent information leakage, proteins with more than 90\% sequence identity to any test protein were excluded, using MMSeqs2 \citep{steinegger2017mmseqs2} with a coverage threshold of 0.8. The number of unique proteins, unique ligands and total data points for each task subset can be found in Table \ref{tab:train_val_datasets}.

\begin{table}[ht]
    \centering
    \caption{Number of unique proteins, unique ligands and \ac{pli} data points in structure-based (SB) and ligand-based (LB) training/validation datasets.}
    
    \label{tab:train_val_datasets}
    \begin{tabular}{lcccccc}
    \toprule
        & \multicolumn{3}{c}{SB Data} & \multicolumn{3}{c}{LB Data} \\
        Task & Proteins & Ligands & Data Points &  Proteins & Ligands & Data Points \\
        \midrule
        Virtual Screening/ Target Fishing & 15,366 & 12,005 & 25,296 & 3,572 & 1,856,963 & 61,468,400 \\
        Pocket Prediction & 4,909 & 2,202 & 16,174 & 3,146 & 1,799,248 & 54,504,697 \\
        Pocket Selection  & 15,509 & 12,273 & 25,511 & 3,566 & 1,858,670 & 62,091,428 \\ \bottomrule
    \end{tabular}
\end{table}

\subsection{Test Datasets}\label{app:sec:test_data}

We evaluated our models on diverse benchmark datasets tailored to each task.  

For virtual screening, we used two widely adopted benchmarks, LIT-PCBA \citep{trannguyen2020litpcba} and DUD-E \citep{mysinger2012directory}.  
The DUD-E dataset contains 22,886 active compounds against 102 protein targets, paired with property-matched decoys designed to mimic physical characteristics of active molecules while differing in topology. LIT-PCBA complements DUD-E by providing experimentally validated high-throughput screening results across 15 targets. Unlike DUD-E, which uses synthetic decoys, LIT-PCBA relies exclusively on assay data, resulting in a more realistic and more challenging benchmark for large-scale virtual screening.

For target fishing, we use the Kinobeads chemical proteomics 
dataset of \citet{reinecke2024chemical}. 
The study profiled 1,183 kinase-directed small molecules in cancer-cell lysates by competitive enrichment on immobilized inhibitors, yielding apparent affinities (Kd\textsuperscript{app}) from a 
two-dose competition design (100 nM and 1 µM) and high-confidence target calls via a trained random-forest classifier. 
We treat these calls as positives and use the remaining measured proteins as negatives when ranking targets per compound.
The raw data are publicly available via ProteomicsDB. After preprocessing and mapping gene symbols to the PDB structure with the highest resolution among those annotated for Homo sapiens (if available), we obtained a dataset of 985 ligands and 2,714 proteins.

Pocket prediction performance was evaluated on three established datasets, which were also used in \citet{sestak2024vnegnn}.  
Coach420 \citep{krivak2018p2rank} is a curated benchmark of 420 proteins with annotated binding sites on single-chain structures. HOLO4K \citep{krivak2018p2rank} consists of over 4,000 holo protein structures with experimentally verified binding pockets, many of which are large multi-chain complexes.
For both, Coach420 and HOLO4K, we adopt the so-called \textit{mlig} subsets, as detailed in \citet{krivak2018p2rank}, which encompass only biologically relevant ligands.
Finally, the PDBbind v.2020 refined set \citep{wang2005pdbbind} includes high-quality protein-ligand complexes with reliable structural and binding affinity data, serving as a stringent benchmark for pocket localization in realistic docking scenarios.

For ligand-conditioned pocket selection, we employed the temporal test split of PDBbind introduced in EquiBind \citep{stark2022equibind}, which ensures temporal separation between training and evaluation complexes, thereby simulating prospective prediction performance. Additionally, we evaluated on the PoseBusters benchmark \citep{buttenschoen2024posebusters}, a curated set of recent high-quality crystal structures designed to test both the physical validity and accuracy of predicted ligand poses. In addition, we constructed a new benchmark based on the Allosteric Site Database (ASD, June 2023 release) \citep{liu2020unraveling}. This dataset comprises protein–ligand complexes annotated with allosteric binding sites, providing a novel and challenging testbed for evaluating the generalization of models beyond orthosteric binding interactions. We filtered out all proteins overlapping with the PDBbind training and validation sets. 

\begin{figure}[h]
    \centering
    \includegraphics[width=\linewidth]{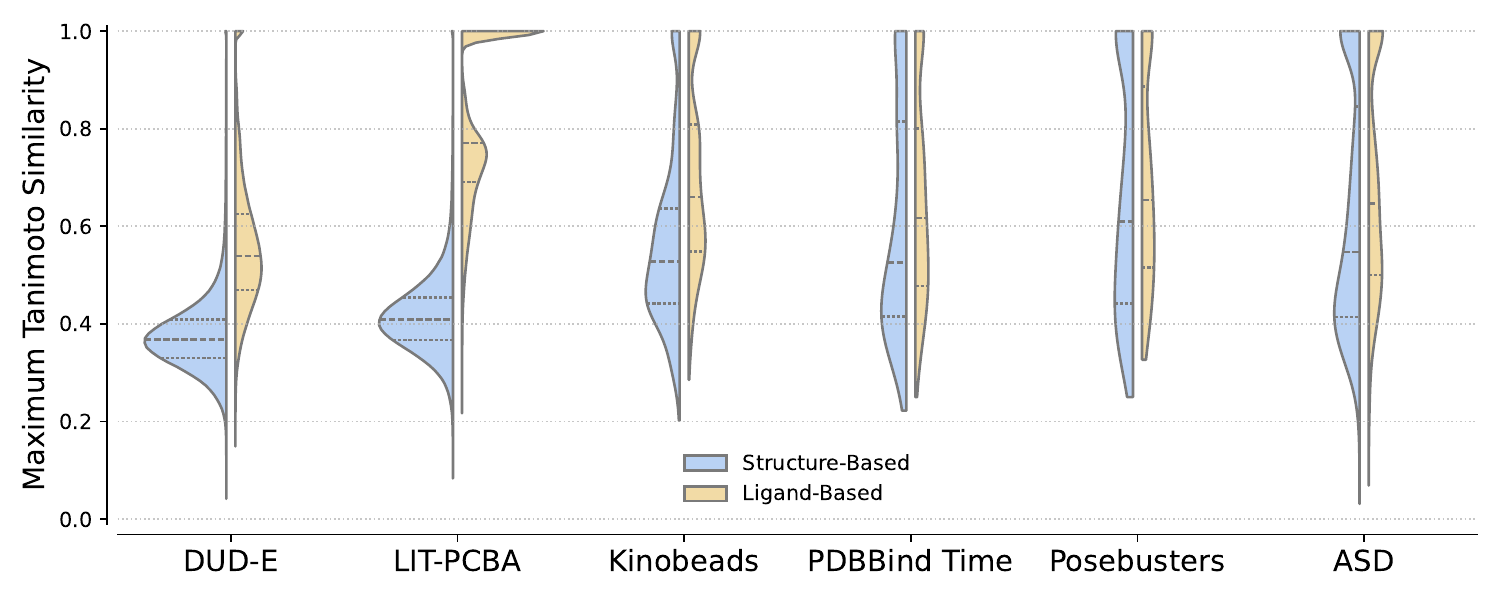}
    \caption{Distributions of maximum Tanimoto similarities between ECFP count fingerprints (radius 2, length 2048) of test-set molecules and those in the structure- and ligand-based training sets.}
    \label{fig:tanimoto_sims}
\end{figure}

Table \ref{tab:test_datasets} summarizes the number of data points, unique proteins and unique ligands for each test dataset. Figure \ref{fig:tanimoto_sims} shows the distribution of maximum ECFP Tanimoto similarities between each test molecule and all training molecules.

\begin{table}
    \centering
    \begin{threeparttable}
        \caption{Summary of test datasets used for evaluation across different tasks. LB = ligand-based datasets, SB = structure-based datasets. \label{tab:test_datasets}}
        
        \begin{tabular}{lcccc}
        \toprule
        Dataset & Type & Data Points & Unique Proteins & Unique Ligands \\
        \midrule 
        DUD-E & LB & 1,434,019 & 102 & 1,200,431 \\
        LIT-PCBA & LB & 2,808,770 & 15 (129)\tnote{a} & 383,772 \\
        Kinobeads & LB & 1,424,686 & 2,714 & 985  \\
        Coach420 & SB & 348 & 300 & 278 \\
        HOLO4K & SB & 4,235 & 3,446 & 1,700 \\
        PDBbind Refined & SB & 5,309 & 5,309 & 4,482 \\
        PDBbind Time & SB & 384 & 321 & 328\\
        ASD & SB & 1802 & 1765 & 1117 \\
        PoseBusters & SB & 274 & 274 & 274 \\
        \bottomrule
        \end{tabular}
    \begin{tablenotes}
        \footnotesize
        \item[a] LIT-PCBA comprises 15 targets, each associated with multiple PDB IDs (129 in total), and during evaluation, predictions are first averaged across all PDB IDs for each target.
    \end{tablenotes} 
\end{threeparttable}
\end{table}

\clearpage
\section{Hyperparameters and Training Details}\label{app:sec:train_details}

For the protein encoder, we adopt VN-EGNN with the default parameters reported by
\citet{sestak2024vnegnn}, i.e., a 5-layer architecture with distinct
weights per layer, input dimension 1280 (from ESM-2 embeddings \citep{lin2023evolutionary}), output dimension 100, SiLU activation, and residual connections. Two linear projection layers are trained to map binding site and protein nodes into the contrastive space of dimension
$D=256$. DBSCAN \citep{ester1996dbscan} clustering on the initial candidate pockets was run with a neighborhood radius of 4 and a minimum cluster size of 1.

Ligands are represented as extended connectivity fingerprints \citep{rogers2010ecfp} with radius 2 and length 2048, concatenated with a vector of 210 chemical descriptors from RdKit \citep{landrum2006rdkit}, yielding an input dimension of 2258. The ligand encoder is a two-layer MLP with hidden dimension 512, output dimension $2D = 512$, GELU activation, 10\% input dropout, and 50\% dropout on the hidden layer.

Training uses a batch size of 64 on structure-based data, resulting in 63 negative
ligands per protein and vice versa through in-batch negative sampling. For ligand-based
training, each batch contains 16 proteins, with actives and inactives sampled at a 1:3
ratio and capped at 10,000 active ligands per protein. Contrastive loss temperature
parameters are set to the inverse square root of the respective embedding dimensions,
i.e., $\tau_{\mathrm{p2m}} = \frac{1}{\sqrt{512}}$ and $\tau_{\mathrm{m2p}} = \tau_{\mathrm{m2b}} = \frac{1}{\sqrt{256}}$. All loss terms are weighted
equally in structure-based training, while the ligand-based loss is scaled by a factor
of 6 to match the magnitude of $L_{SB}$.

We optimize using AdamW \citep{loshchilov2017decoupled} with an initial learning rate of $10^{-3}$. A learning rate
scheduler reduces the rate by a factor of 10 when the validation metric does not
improve for 30 epochs, with a minimum learning rate of $10^{-6}$. Early stopping with
a patience of 100 epochs is applied based on the same validation metric. Separate models
were trained for each task due to the different data splits and training was conducted on NVIDIA A100 GPUs with 40GB memory for 200--350 epochs.

\section{Metrics}\label{app:sec:metrics}

Depending on the task, we employ different evaluation metrics, which are described below.

For virtual screening, we evaluate the area under the receiver operating characteristic curve (AUROC), the Boltzmann-enhanced discrimination of ROC (BEDROC) at $\alpha=85$, and enrichment factors (EF) at the top 0.5\%, 1\% and 5\%. Unlike AUROC, which treats all parts of the ranking equally and is therefore a strong general-purpose metric, BEDROC is tailored to virtual screening scenarios where early recognition of actives is critical \citep{truchon2007early}. The EF at top $x\%$ quantifies the over-representation of actives among the highest-ranked molecules. An EF of 1 corresponds to random ranking, while larger values indicate stronger enrichment.

For target fishing, we evaluate $\Delta$AUPRC, which measures the improvement in area under the precision–recall curve relative to a random baseline, in addition to AUROC and EF at 1\%, 

For binding pocket prediction, we measure the DCC (distance from predicted pocket center to ground-truth pocket center) or DCA (distance from predicted pocket center to the closest atom of the corresponding ligand) success rates at 4\r{A}. For a protein with $k$ ground-truth pockets, we consider the $k$ top-ranked binding sites. The success rate is the fraction of ground-truth pockets where at least one predicted pocket satisfies the DCC or DCA threshold of 4\r{A}.

For ligand-conditioned pocket selection, we consider the DCC success rate at 4\r{A} of the top-ranked predicted pocket compared to all ground-truth pockets associated with the given ligand.

\clearpage
\section{Ablation Studies}\label{app:sec:ablation}

We performed an ablation study on the main components of ConGLUDe, including structure- and ligand-based training data as well as the different geometric and contrastive loss terms. The results are shown in Tables~\ref{tab:virtual_screening_ablation}--\ref{tab:pocket_selection_ablation}.

On the virtual screening dataset LIT-PCBA, ablating any component leads to a deterioration of the performance metrics, indicating that all components together contribute to the effectiveness of ConGLUDe. On DUD-E, structure-based data are critical, whereas ligand-based data are not strictly necessary and their removal even slightly improves performance. We do not interpret this as evidence against the benefits of unified learning: the same unified model shows clear positive transfer on the more realistic LIT-PCBA benchmark, and DUD-E is widely recognized to be substantially biased \citep{chen2019hidden}, which is why we restrict its discussion to the appendix.

\begin{figure}[ht]
  \centering
  \includegraphics[width=\textwidth]{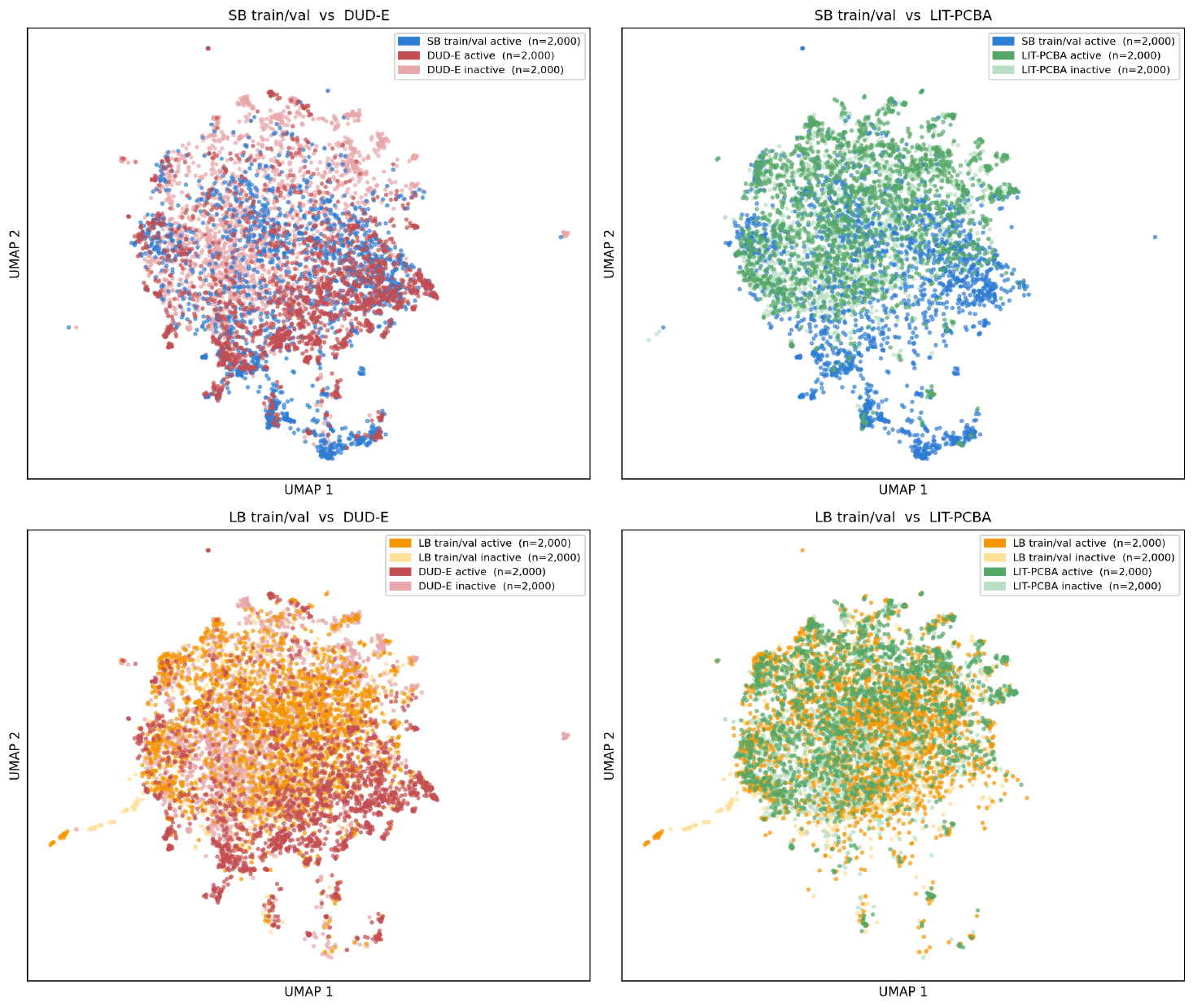}
  \caption{UMAP projections of Morgan fingerprints of randomly sampled structure-based (SB) and ligand-based (LB) training/validation molecules and test set molecules from DUD-E and LIT-PCBA split into actives and inactives/decoys.}
  \label{fig:umap}
\end{figure}

We hypothesize that the performance drop on DUD-E stems from a chemical space mismatch between the ligand-based training data and DUD-E. A UMAP visualization of molecular fingerprints (Figure~\ref{fig:umap}) reveals two contributing factors: first, DUD-E actives and decoys form distinct clusters in chemical space, confirming that this benchmark is substantially easier than real-world screening. Second, DUD-E actives occupy similar regions to structure-based training molecules, while covering a more disjoint region relative to ligand-based training data. LIT-PCBA, by contrast, shows a distribution closely aligned with the ligand-based training data, consistent with the positive transfer observed there. Together, this reinforces the consensus that DUD-E's artificially constructed decoys do not reflect realistic screening conditions, making it poorly suited for benchmarking virtual screening methods.

Ablating individual loss terms not directly tied to virtual screening has little effect on DUD-E. The slight improvement under the "no $\mathcal{L}_{\mathrm{geometric}}$" setting may be attributed to additional model capacity being available for virtual screening when pocket prediction does not need to be learned. However, performance drops on LIT-PCBA, confirming that this loss term does not hurt generalization.

\begin{table*}[h]
    \centering
    \begin{threeparttable}
        \caption{Ablation study: Performance on virtual screening on the DUD-E \citep{mysinger2012directory} and LIT-PCBA \citep{trannguyen2020litpcba} datasets measured by AUROC, BEDROC and EF at 1\%. ConGLUDe results are reported from a single run using the same random seed as the ablated models.}
        \label{tab:virtual_screening_ablation}
        \footnotesize
        \begin{tabular}{
            >{\raggedright\arraybackslash}p{0.18\textwidth}
            >{\centering\arraybackslash}p{0.1\textwidth}
            >{\centering\arraybackslash}p{0.1\textwidth}
            >{\centering\arraybackslash}p{0.1\textwidth}
            >{\centering\arraybackslash}p{0.1\textwidth}
            >{\centering\arraybackslash}p{0.1\textwidth}
            >{\centering\arraybackslash}p{0.1\textwidth}}
            \toprule
             & \multicolumn{3}{c}{DUD-E} & \multicolumn{3}{c}{LIT-PCBA} \\
             \cmidrule(lr){2-4} \cmidrule(lr){5-7}
             & AUROC$\uparrow$ & BEDROC$\uparrow$ & EF $1 \%\uparrow$ & AUROC$\uparrow$ & BEDROC$\uparrow$ & EF $1 \%\uparrow$ \\
            \midrule
            only SB data                 & 83.88 & 56.20 & 36.57 & 53.06 & 5.48  & 4.73 \\
            only LB data                 & 67.11 & 10.61 & 5.31  & 67.94 & 11.11 & 9.38 \\
            \midrule
            no $\mathcal{L}_{\mathrm{geometric}}$ & 83.26 & 53.05 & 34.79 & 64.17 & 11.41 & 10.06 \\
            no $\mathcal{L}_{\mathrm{m2p}}$       & 82.58 & 50.30 & 32.26 & 64.80 & 11.01 & 10.24 \\
            no $\mathcal{L}_{\mathrm{m2b}}$       & 81.55 & 50.16 & 32.34 & 64.90 & 10.97 & 8.98 \\
            \midrule
            ConGLUDe & 82.04 & 50.80 & 32.52 & 66.25 & 13.63 & 12.25 \\
            \bottomrule
        \end{tabular}
    \end{threeparttable}
\end{table*}

For target fishing (Table~\ref{tab:target_fishing_ablation}), omitting structure-based training data leads to a near-complete collapse in performance, underscoring its importance for learning meaningful protein representations. Contrary to what one might expect, $\mathcal{L}_{\mathrm{m2p}}$ does not appear to play a significant role for target fishing, but shows benefits for virtual screening and pocket selection.

\begin{table*}[h]
    \centering
    \begin{threeparttable}
        \caption{Ablation study: Performance on target fishing on the Kinobeads datasets measured by mean AUROC, $\Delta$AUPRC, and EF at 1\% averaged across small molecules.}
        \label{tab:target_fishing_ablation}
        \footnotesize
        \begin{tabular}{
            >{\raggedright\arraybackslash}p{0.18\textwidth}
            >{\centering\arraybackslash}p{0.1\textwidth}
            >{\centering\arraybackslash}p{0.1\textwidth}
            >{\centering\arraybackslash}p{0.1\textwidth}
            >{\centering\arraybackslash}p{0.1\textwidth}
            >{\centering\arraybackslash}p{0.1\textwidth}
            >{\centering\arraybackslash}p{0.1\textwidth}}
            \toprule
             & AUROC $\uparrow$ & $\Delta$AUPRC $\uparrow$ & EF $1 \%\uparrow$ \\
            \midrule
            only SB data                 & 63.2 & 6.4 & 11.4 \\
            only LB data                 & 45.2 & 0.3 & 0.5 \\
            \midrule
            no $\mathcal{L}_{\mathrm{geometric}}$ & 63.5 & 6.1 & 11.5 \\
            no $\mathcal{L}_{\mathrm{m2p}}$       & 65.4 & 4.7 & 9.6 \\
            no $\mathcal{L}_{\mathrm{m2b}}$       & 63.5 & 5.3 & 9.6  \\
            \midrule
            ConGLUDe & 65.6 & 5.1 & 9.9 \\
            \bottomrule
        \end{tabular}
    \end{threeparttable}
\end{table*}

For ligand-conditioned pocket selection (Table~\ref{tab:pocket_selection_ablation}), the ablations clearly show that $\mathcal{L}_{\mathrm{geometric}}$ and $\mathcal{L}_{\mathrm{m2b}}$ are essential: omitting the former renders the model unable to identify correct binding sites, while removing the latter significantly diminishes pocket ranking performance. The remaining components have little impact on pocket selection across datasets, confirming that they are not directly relevant to this task but do not interfere with it either.

\begin{table}[h]
    \centering
    \caption{Performance of ligand-conditioned pocket selection measured by the top-1 DCC success rate at a 4\AA\ threshold ($\uparrow$). Values in parentheses indicate 95\% confidence intervals. DC = DrugCLIP.}
    \label{tab:pocket_selection_ablation}
    \footnotesize
    \begin{threeparttable}
        \begin{tabular}{lccc}
            \toprule
             & PDBbind Time & PoseBusters & ASD \\
            \midrule
            only SB data & 0.55$^{(.50,.60)}$ & 0.25$^{(.23,.28)}$ & 0.29$^{(.27,.31)}$  \\
            only $\mathcal{L}_{\mathrm{geometric}}$ and $\mathcal{L}_{\mathrm{m2b}}$ & 0.57$^{(.52,.62)}$ & 0.27$^{(.24,.29)}$ & 0.31$^{(.29,.32)}$\\
            \midrule
            no $\mathcal{L}_{\mathrm{geometric}}$ & 0.04$^{(.03,.07)}$ & 0.00$^{(.00,.01)}$ & 0.01$^{(.00,.01)}$ \\
            no $\mathcal{L}_{\mathrm{m2b}}$ & 0.35$^{(.31,.40)}$ & 0.11$^{(.09,.13)}$ & 0.10$^{(.09,.12)}$ \\
            no $\mathcal{L}_{\mathrm{m2p}}$ & 0.47$^{(.42,.52)}$ & 0.22$^{(.19,.24)}$ & 0.24$^{(.22,.25)}$ \\
            \midrule
            ConGLUDe & 0.54$^{(.47,.60)}$ & 0.27$^{(.24,.31)}$ & 0.35$^{(.33,.37)}$\\
            \bottomrule
        \end{tabular}
    \end{threeparttable}
\end{table}

\section{Extended Results}\label{app:sec:extended_results}

\subsection{Virtual Screening}\label{app:sec:vs}
Tables \ref{tab:lit_pcba} and \ref{tab:dude} show the complete evaluation on LIT-PCBA and DUD-E benchmarks, respectively. We compare against a diverse set of representative baselines, detailed below, spanning classical docking engines, pose-aware machine learning models, and modern contrastive learning methods.

Classical docking methods such as Glide-SP~\citep{halgren2004glide}, AutoDock Vina~\citep{trott2010vina}, and Surflex~\citep{spitzer2012surflexdock} explicitly sample ligand conformations within predefined binding pockets and rank poses using empirical or physics-inspired scoring functions. Pose-based machine learning methods, including NN-Score~\citep{durrant2011nnscore2}, RF-Score~\citep{ballester2010rfscore}, and Gnina~\citep{mcnutt2021gnina}, operate on docked complexes and leverage learned scoring functions to predict binding affinity or pose quality.

Deep learning models that require explicit pocket definitions are also included in the comparison. These comprise voxel-based 3D CNNs such as Pafnucy~\citep{stepniewska2017pafnucy}, distance-based representations such as OnionNet~\citep{zheng2019onionnet}, and graph neural network approaches including BigBind~\citep{brocidiacono2024bigbind} and PLANET~\citep{zhang2024planet}. These methods directly exploit geometric and chemical features of the binding site and are typically applied as docking rescoring or refinement tools.

More recent contrastive and embedding-based approaches aim to bypass explicit docking by learning joint representations of proteins and ligands. DrugCLIP~\citep{gao2023drugclip}, DrugHash~\citep{han2024hashing}, S$^2$Drug~\citep{he2025s2drug}, LigUnity~\citep{feng2025hierarchical}, and HypSeek~\citep{wang2025hypseek} align ligand embeddings with structure-based pocket representations, enabling fast similarity-based virtual screening. In contrast, SPRINT~\citep{mcnutt2024sprint} uses whole-protein representations and is trained on ligand-based datasets, which supports broad target coverage but does not allow pocket-specific predictions.

On the DUD-E dataset, methods that use the known binding pocket as input significantly outperform approaches that do not assume pocket information, like ConGLUDe. However, when the binding pocket is unknown, which is common in many virtual screening settings, pocket-based methods would need to rely on predicted binding sites. In this scenario, the performance of pocket-based methods drops substantially and becomes markedly worse than that of ConGLUDe. Notably, the other natively pocket agnostic method, SPRINT, also performs significantly worse than ConGLUDe on DUD-E.

On the more realistic LIT-PCBA benchmark, natively pocket-agnostic approaches significantly outperform pocket-requiring methods, even when the correct pocket is provided as input to the latter. Overall, ConGLUDe is the only model that demonstrates strong cross-benchmark generalization.

\begin{figure}[ht]
  \centering
  \includegraphics[width=0.6\textwidth]{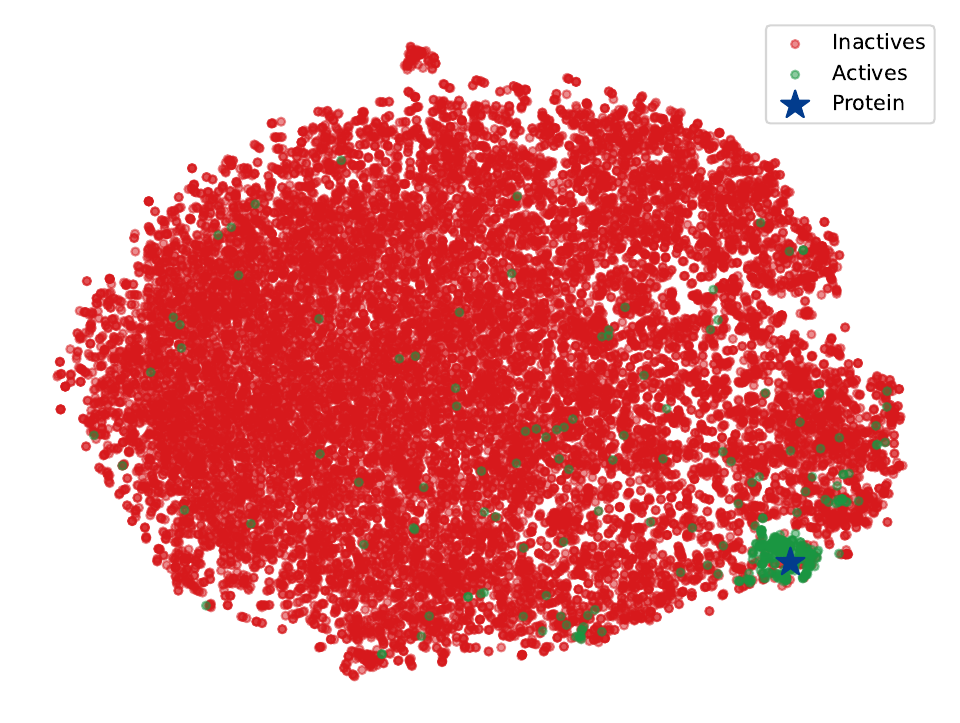}
  \caption{t-SNE projection of protein and ligand embeddings for the DUD-E target with PDB ID 2FSZ.}

  \label{fig:tsne}
\end{figure}

To visualize the learned representation space, we applied t-SNE to project both protein and ligand embeddings into two dimensions. As shown by one example in Figure~\ref{fig:tsne}, active ligands cluster around the embedding of their target protein, whereas inactive molecules are distributed more diffusely across the space. This pattern highlights the model’s ability to capture meaningful protein–ligand relationships.

\begin{table}[h]
    \centering
        \caption{Zero-shot virtual screening performance on the LIT-PCBA dataset measured by AUROC, BEDROC, and EF at 0.5\%, 1\% and 5\%.
        For ConGLUDe, we report the median and mean absolute deviation (MAD) over three training re-runs. The best value per column as well as all other values within one MAD are marked in bold.}
        \label{tab:lit_pcba}
        \begin{tabular}{
            >{\raggedright\arraybackslash}p{0.18\textwidth}
            >{\centering\arraybackslash}p{0.13\textwidth}
            >{\centering\arraybackslash}p{0.13\textwidth}
            >{\centering\arraybackslash}p{0.13\textwidth}
            >{\centering\arraybackslash}p{0.13\textwidth}
            >{\centering\arraybackslash}p{0.13\textwidth}}
        \toprule
         & \multirow{2}{*}{ AUROC (\%) } & {\multirow{2}{*} {BEDROC (\%)}} & \multicolumn{3}{c}{ EF } \\
        & & & $0.5 \%$ & $1 \%$ & $5 \%$ \\
        \midrule 
        Surflex & 51.47 & - & - & 2.50 & - \\
        Glide-SP & 53.15 & 4.00 & 3.17 & 3.41 & 2.01 \\
        Planet & 57.31 & - & 4.64 & 3.87 & 2.43 \\
        GninA & 60.93 & 5.40 & - & 4.63 & - \\
        DeepDTA & 56.27 & 2.53 & - & 1.47 & - \\
        BigBind & 60.80 & - & - & 3.82 & - \\
        \midrule
        DrugCLIP & 57.17 & 6.23 & 8.56 & 5.51 & 2.27 \\
        DrugHash & 54.58 & 7.14 & 9.65 & 6.14 & 2.42 \\
    	S$^2$Drug & 58.23 & 8.69 & 11.44 & 7.38 & 2.97 \\
        LigUnity & 59.85 & \textbf{11.33} & - & 6.47 & - \\
        HypSeek & 62.10 & \textbf{11.96} & - & 6.81 & - \\
        \midrule
        DrugCLIP\textsubscript{P2Rank} & 51.23 & 2.65 & 2.44 & 1.80 & 1.36 \\
        DrugCLIP\textsubscript{VN-EGNN} & 56.70 & 4.15 & 2.51 & 3.17 & 2.04 \\
        SPRINT & \textbf{73.40} & \textbf{ 12.30} & \textbf{15.90} & \textbf{10.78} & \textbf{5.29} \\
        \midrule
        ConGLUDe & 64.06 $^{\pm 3.25}$ & \textbf{12.24} $^{\pm2.06}$ & \textbf{15.87} $^{\pm2.03}$ & \textbf{ 11.03 $^{\pm1.81}$} & \textbf{4.68} $^{\pm0.30}$ \\ 
        \bottomrule
        \end{tabular}

\end{table}

\begin{table}[h]
    \centering
        \caption{Zero-shot virtual screening performance on the DUD-E dataset measured by AUROC, BEDROC, and EF at 0.5\%, 1\% and 5\%.
        For ConGLUDe, we report the median and mean absolute deviation (MAD) over three training re-runs. The best value per column as well as all other values within one MAD are marked in bold.}
        \label{tab:dude}
                \begin{tabular}{
            >{\raggedright\arraybackslash}p{0.18\textwidth}
            >{\centering\arraybackslash}p{0.13\textwidth}
            >{\centering\arraybackslash}p{0.13\textwidth}
            >{\centering\arraybackslash}p{0.13\textwidth}
            >{\centering\arraybackslash}p{0.13\textwidth}
            >{\centering\arraybackslash}p{0.13\textwidth}}
        \toprule
        & \multirow{2}{*}{ AUROC (\%) } & {\multirow{2}{*} {BEDROC (\%)}} & \multicolumn{3}{c}{ EF } \\
        & & & $0.5 \%$ & $1 \%$ & $5 \%$ \\
        \midrule 
        Glide-SP & 76.70 & 40.70 & 19.39 & 16.18 & 7.23 \\
        Vina & 71.60 & - & 9.13 & 7.32 & 4.44 \\
        NN-score & 68.30 & 12.20 & 4.16 & 4.02 & 3.12 \\
        RFscore & 65.21 & 12.41 & 4.90 & 4.52 & 2.98 \\
        Pafnucy & 63.11 & 16.50 & 4.24 & 3.86 & 3.76 \\
        OnionNet & 59.71 & 8.62 & 2.84 & 2.84 & 2.20 \\
        Planet & 71.60 & - & 10.23 & 8.83 & 5.40 \\
        \midrule
        DrugCLIP & 80.93 & 50.52 & 38.07 & 31.89 & 10.66 \\
        DrugHash & 83.73 & 57.16 & 43.03 & 37.18 & 12.07 \\
        S$^2$Drug & 92.46 &\textbf{ 79.25} & \textbf{58.37} & 43.06 & \textbf{18.82} \\
        LigUnity & 89.22 & 65.26 & - & 42.63 & - \\
        HypSeek &  \textbf{94.35} & 78.92 & - & \textbf{51.44} & - \\
        \midrule
        DrugCLIP\textsubscript{P2Rank} & 62.70 & 7.00 & 4.02 & 3.73 & 2.42 \\
        DrugCLIP\textsubscript{VN-EGNN} & 71.08 & 28.22 & 20.47 & 16.91 & 6.93 \\        
        \midrule
        ConGLUDe & 81.29 $^{\pm 1.11}$ & 49.49 $^{\pm 1.94}$ & 39.43 $^{\pm 0.97}$ &  31.76 $^{\pm 1.13}$ & 10.71 $^{\pm 0.26}$ \\ 
        \bottomrule
        \end{tabular}        
\end{table}

\clearpage
\subsection{Binding Site Prediction}
Table~\ref{tab:app_pp_results} reports binding site identification results for all methods compared in \citet{sestak2024vnegnn}.

\begin{table}[ht]
    \centering
    \begin{threeparttable}
        \caption{Performance at binding site identification in 
        terms of DCC and DCA success rates at 4\r{A} on the COACH420, HOLO4K and PDBbind2020 refined datasets. The best performing method per column is marked bold, the second best in italics.\\}
        \label{tab:app_pp_results}
        \begin{tabular}{
            >{\raggedright\arraybackslash}p{0.18\textwidth}
            >{\centering\arraybackslash}p{0.1\textwidth}
            >{\centering\arraybackslash}p{0.1\textwidth}
            >{\centering\arraybackslash}p{0.1\textwidth}
            >{\centering\arraybackslash}p{0.1\textwidth}
            >{\centering\arraybackslash}p{0.1\textwidth}
            >{\centering\arraybackslash}p{0.1\textwidth}}
            \toprule
            \multirow{2}{*}{Methods}
            & \multicolumn{2}{c}{COACH420}
            & \multicolumn{2}{c}{HOLO4K}
            & \multicolumn{2}{c}{PDBbind2020} \\
            \cmidrule(lr){2-3}\cmidrule(lr){4-5}\cmidrule(lr){6-7}
            & DCC$\uparrow$ & DCA$\uparrow$
            & DCC$\uparrow$ & DCA$\uparrow$
            & DCC$\uparrow$ & DCA$\uparrow$ \\
            \midrule
            Fpocket       & 0.228 & 0.444 & 0.192 & 0.457 & 0.253 & 0.371 \\
            P2Rank        & 0.464 & \emph{0.728} & 0.474 & \textbf{0.787} & 0.653 & \textit{0.826} \\
            \midrule
            DeepSite      & --    & 0.564 & --    & 0.456 & --    & --    \\
            Kalasanty     & 0.335 & 0.636 & 0.244 & 0.515 & 0.416 & 0.625 \\
            DeepSurf      & 0.386 & 0.658 & 0.289 & 0.635 & 0.510 & 0.708 \\
            DeepPocket    & 0.399 & 0.645 & 0.456 & 0.734 & 0.644 & 0.813 \\
            \midrule
            GAT           & 0.039 & 0.130 & 0.036 & 0.110 & 0.032 & 0.088 \\
            GCN           & 0.049 & 0.139 & 0.044 & 0.174 & 0.018 & 0.070 \\
            GAT + GCN     & 0.036 & 0.131 & 0.042 & 0.152 & 0.022 & 0.074 \\
            GCN2          & 0.042 & 0.131 & 0.051 & 0.163 & 0.023 & 0.089 \\
            \midrule
            SchNet        & 0.168 & 0.444 & 0.192 & 0.501 & 0.263 & 0.457 \\
            EGNN          & 0.156 & 0.361 & 0.127 & 0.406 & 0.143 & 0.302 \\
            EquiPocket    & 0.423 & 0.656 & 0.337 & 0.662 & 0.545 & 0.721 \\
            \midrule
            VN-EGNN       & \textbf{0.605} & \textbf{0.750} & \textbf{0.532} & 0.659 & \emph{0.669} & 0.820 \\
            \midrule
            ConGLUDe   & \emph{0.602} & 0.726 & \emph{0.525} & \emph{0.693} & \textbf{0.689} & \textbf{0.856} \\
            \bottomrule
        \end{tabular}
            
    \end{threeparttable}
\end{table}


\end{document}

%% file: math_commands.tex
\usepackage{amsmath,amsfonts,bm}

\def\eqref#1{equation~\ref{#1}}

\def\1{\bm{1}}

\DeclareMathAlphabet{\mathsfit}{\encodingdefault}{\sfdefault}{m}{sl}
\SetMathAlphabet{\mathsfit}{bold}{\encodingdefault}{\sfdefault}{bx}{n}

